\RequirePackage{fix-cm}
\documentclass[twocolumn]{svjour3}          
\smartqed  
\usepackage{graphicx}

\usepackage{url}
\usepackage{array}
\usepackage{subfigure}
\usepackage{amsmath}
\usepackage{float}
\usepackage{color}
\usepackage{amsmath}

\usepackage{booktabs}
\usepackage{multirow} 
\usepackage[breaklinks=true,colorlinks,citecolor=blue,urlcolor=blue,linkcolor=blue,bookmarks=false,pagebackref=true]{hyperref}
\usepackage{amssymb}

\newcolumntype{M}[1]{>{\centering\arraybackslash}m{#1}}

\begin{document}

\title{Beyond Monocular Deraining: Parallel Stereo Deraining Network Via Semantic Prior}

\author{Kaihao Zhang         \and
        Wenhan Luo        \and
        Yanjiang Yu       \and
        Wenqi Ren        \and 
        Fang Zhao        \and \\
        Changsheng Li  \and 
        Lin Ma \and
        Wei Liu  \and 
        Hongdong Li
}

\institute{Kaihao Zhang and Hongdong Li \at
        Australian National University, Australia\\
        \email{\{kaihao.zhang, hongdong.li\}@anu.edu.au}        \\
        \and
         Wenhan Luo and Wei Liu\at
         Tencent, Shenzhen, China \\
          \email{\{whluo.china@gmail.com; wl2223@columbia.edu\}}      \\
          \and
        Yanjiang Yu and Changsheng Li \at
        Beijing Institute of Technology , Beijing, China \\
        \email{\{yuyanjiang87@gmail.com; lcs@bit.edu.cn\}} \\
          \and
          Wenqi Ren \at
         Institute of Information Engineering, Chinese Academy of Sciences, Beijing, China\\
        \email{renwenqi@iie.ac.cn} \\
        \and
         Fang Zhao \at
         Inception Institute of Artificial Intelligence, Abu Dhabi, UAE\\
        \email{fang.zhao@inceptioniai.org} \\
         \and
         Lin Ma \at
         Meituan Group, Beijing, China \\
           \email{forest.linma@gmail.com} \\
}

\date{Received: date / Accepted: date}

\maketitle

\begin{abstract}
Rain is a common natural phenomenon. Taking images in the rain however often results in degraded quality of images, thus compromises the performance of many computer vision systems. Most existing de-rain algorithms use only one single input image and aim to recover a clean image. Few work has exploited stereo images. Moreover, even for single image based monocular deraining, many current methods fail to complete the task satisfactorily because they mostly rely on per pixel loss functions and ignore semantic information. In this paper, we present a Paired Rain Removal Network (PRRNet), which exploits both stereo images and semantic information. Specifically, we develop a Semantic-Aware Deraining Module (SADM) which solves both tasks of semantic segmentation and deraining of scenes, and a Semantic-Fusion Network (SFNet) and a View-Fusion Network (VFNet) which fuse semantic information and multi-view information respectively. In addition, we also introduce an Enhanced Paired Rain Removal Network (EPRRNet) which exploits semantic prior to remove rain streaks from stereo images. We first use a coarse deraining network to reduce the rain streaks on the input images, and then adopt a pre-trained semantic segmentation network to extract semantic features from the coarse derained image. Finally, a parallel stereo deraining network fuses semantic and multi-view information to restore finer results. We also propose new stereo based rainy datasets for benchmarking. Experiments on both monocular and the newly proposed stereo rainy datasets demonstrate that the proposed method achieves the state-of-the-art performance.
\end{abstract}

\section{Introduction}

Stereo image processing has become an increasingly active research field in computer vision with the development of stereoscopic vision. Based on stereo images, many key technologies such as depth estimation \cite{godard2017unsupervised,liu2015deep,riegler2019connecting}, scene understanding \cite{eslami2016attend,shao2015deeply,zhao2017pyramid} and stereo matching \cite{luo2016efficient,chang2018pyramid,pang2017cascade} have achieved a great success. As a common natural phenomenon, rain causes visual discomfort and degrades the quality of images, which can deteriorate the performance of many core models in outdoor vision-based systems. However, there are few studies for stereo deraining. In this paper, we address the problem of removing rain from stereo images.

\begin{figure}[t] 
\label{fig:idea:a} 
  \centering
  {\includegraphics[width=0.99\linewidth]{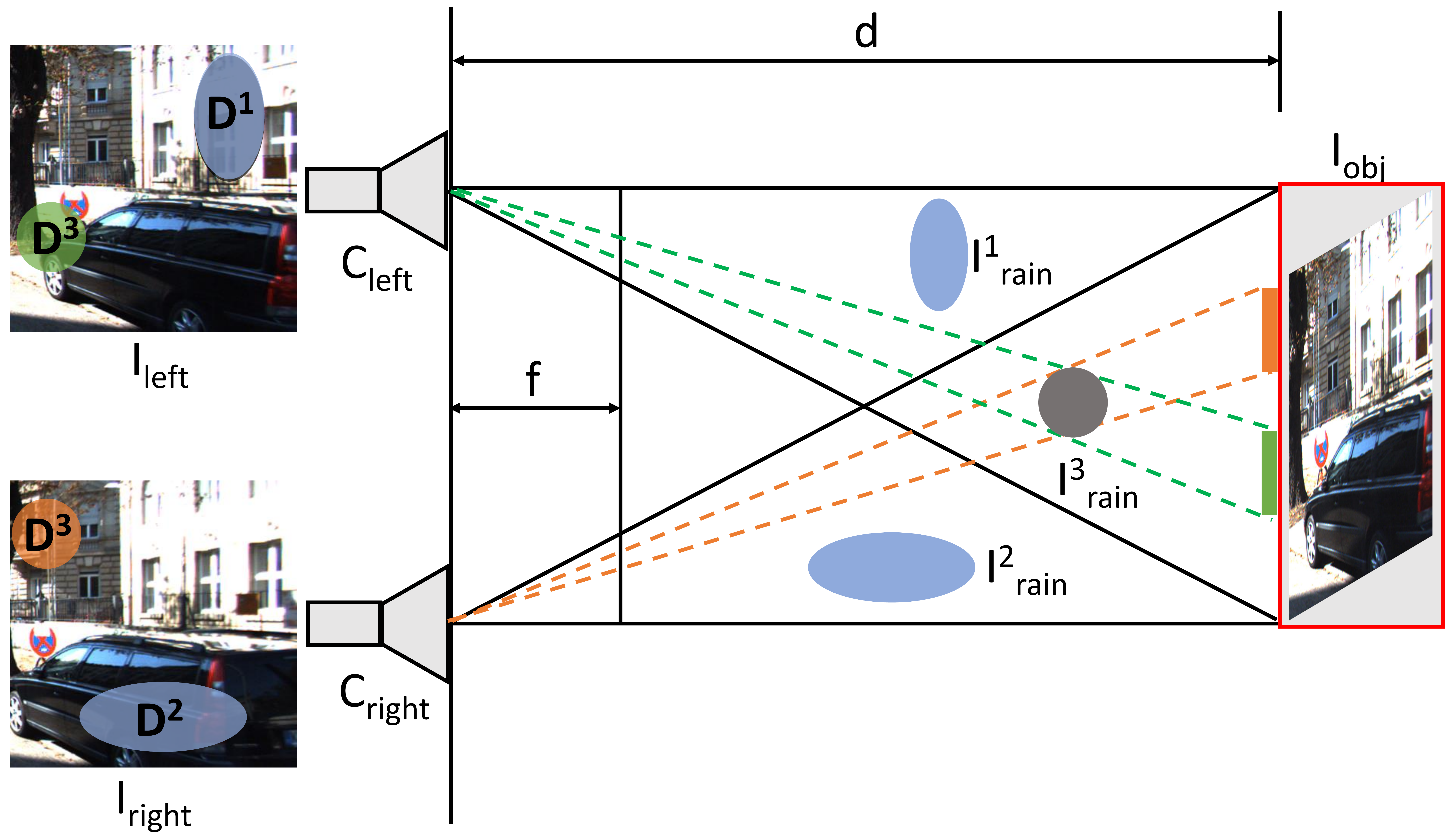}}
  \caption{{\bf The illustration of stereo cameras}. One pair of images $I_{left}$ and $I_{right}$ is captured by stereo cameras $C_{left}$ and $C_{right}$. $I_{obj}$ is an object and $I^{ref}_{obj}$ is the reflection of $I_{obj}$. $d$ is distance between the object and the camera. $f$ is the camera focal length. The same rain $I^3_{rain}$ can cause different effects on images from two views.} 
\end{figure}

In fact, stereo deraining has an intrinsic advantage over monocular deraining because the effects of identical rain streaks in corresponding pixels from stereo images are different. As Fig. \ref{fig:idea:a} shows, the mapping of object $I_{obj}$ on stereo images $I_{left}$ and $I_{right}$ can be represented as
\begin{equation}
I_{left} = I_{obj} * \frac{d}{f} ,
\end{equation}
\begin{equation}
I_{right} = I^{ref}_{obj} * \frac{d}{f} ,
\end{equation}
where $d$ and $f$ are the distance between object and camera and the camera focal length, respectively. $I^{ref}_{obj}$ is the reflection of $I_{obj}$. Assuming that the object $I_{obj}$ is in the middle of two cameras, the lengths of identical objects, $I^{ref}_{obj}$ and $I_{obj}$, on stereo views are the same. However, the effects of rain across stereo images are different. For example, the degraded regions by rain $I^1_{rain}$ on the two images can be denoted as
\begin{equation}
D^1_{left} = I^1_{rain} * \frac{d_{rain}}{f} , 
\end{equation}
\begin{equation}
D^1_{right} = 0 ,
\end{equation}
$I^1_{rain}$ degrades the quality of the object on the left image but does not affect the visual comfort of the right view. $d_{rain}$ is the distance between the camera and the raindrop. There is also rain influencing different regions on both stereo images like $I^3_{rain}$. The image in Fig. \ref{fig:idea:a} demonstrates the different effects of identical rain streaks on stereo views.

Moreover, the geometric cue and semantics provide important prior information, serving as a latent advantage for removing rain. Recently, most deep monocular deraining methods achieve a great success by reconstructing objects based on pixel-level objective functions like MSE. However, these methods ignore modeling the geometric structure of objects and understanding the semantic information of scenes, which in fact benefit deraining.
Hu \textit{et al.} \cite{hu2019depth} try to remove rain via depth estimation, but they fail to understand the rainy scenes.

In this paper, we first propose a semantic-aware deraining module, \textit{SADM}, which removes rain by leveraging scene understanding. Fig. \ref{semantic} illustrates the concept of \textit{SADM}. It contains two parts. The first part is an encoder which takes a rainy image as input and encodes it as semantic-aware features. Then the representations are fed into the second part, a conditional generator, to transform them into the deraining image and scene segmentation. Based on a multi-task shared learning mechanism and different input conditions, the single \textit{SADM} is capable of jointly removing rain and understanding scenes. To further enhance the understanding of input images, a \textit{Semantic-Rethinking Loop} is proposed to utilize the difference between the outputs of the conditional generators in different stages.

Based on \textit{SADM}, we then present a stereo deraining model, \textit{Paired Rain Removal Network (PRRNet)}, which consists of \textit{SADM}, \textit{Semantic-Fusion Network (SFNet)} and \textit{View-Fusion Network (VFNet)}. \textit{SADM} is utilized to learn the semantic information and reconstruct deraining images, while \textit{SFNet} and \textit{VFNet} are to fuse the semantic information with coarse deraining images, and obtain the final deraining images by fusing stereo views, respectively.

Considering that there exist semantic segmentation models which are trained on clean images. We also propose an \textit{Enhanced Paired Rain Removal Network (EPRRNet)} to remove rain via utilizing the semantic prior extracted from pre-trained semantic segmentation networks. The \textit{EPRRNet} consists of three sub-networks: a coarse deraining network, a pre-trained semantic segmentation network and a parallel stereo deraining network. The coarse deraining network first reduces rain streaks from input rainy images and generates derained results. The pre-trained semantic segmentation network then estimates the semantic labels from the derained images. Finally, the derained images and semantic labels are fed into a parallel multi-scale deraining network to generate the final derained images via extracting semantic-attentional features.

Our proposed \textit{EPRRNet} is a variant of the \textit{PRRNet} with a number of critical extensions: (1) We directly use a pre-trained semantic network to extract semantic labels as prior. In this way, our \textit{EPRRNet} is flexible to use up-to-date semantic segmentation networks to help remove rain streak. Meanwhile, the coarse deraining network can alleviate the negative effect of rain for semantic segmentation. (2) A multi-scale derained network is proposed to extract multi-scale features to obtain better details. (3) We use a new parallel cooperation way to fuse stereo images which can make better use of stereo information from low-level to high-level layers.

Currently, there is no public large-scale stereo rainy datasets. In order to evaluate the performance of the proposed method and compare against the state-of-the-art methods, two large stereo rainy image datasets are thus constructed.

In summary, the contributions of this paper are four-fold:
\begin{itemize}
\item
Firstly, a multi-task shared learning deraining model, \textit{SADM}, is proposed to remove rain via scene understanding. This model not only considers pixel-level objective functions like previous methods, but also models the geometric structure and semantic information of input rainy images. Inside \textit{SADM}, a novel \textit{Semantic-Rethinking Loop} is employed to further strengthen the connection between scene understanding and image deraining. 
\item
Secondly, we propose \textit{PRRNet}, the first semantic-aware stereo deraining network. \textit{PRRNet} fuses the semantic information and multi-view information via \textit{SFNet} and \textit{VFNet}, respectively, to obtain the final stereo deraining images. 

\item
Thirdly, we extend the \textit{PRRNet} to an enhanced version, \textit{EPRRNet}. It is a multi-scale network which can flexibly use up-to-date semantic segmentation model to obtain semantic prior. Meanwhile, it also use a parallel cooperation way to fuse stereo information from low-level to high-level layers. 

\item
Finally, we synthesize two stereo rainy datasets for stereo deraining, which may be the largest datasets for stereo image deraining. Experiments on the monocular and stereo rainy datasets show that the proposed \textit{PRRNet} and \textit{EPRRNet} achieve the state-of-the-art performance on both monocular and stereo deraining.
\end{itemize}

\section{Related Work}


\subsection{Single Image Deraining}

Deraining from a single rainy image is a highly ill-posed task, whose mathematical formulation is expressed as
\begin{equation}
O = B + R\,,
\end{equation}
where $O$, $B$ and $R$ are the observed rainy image, the latent clean image and the rain-streak component, respectively.

For traditional methods of recovering the clean deraining image $B$ from the rainy version $O$, Kang \textit{et al.} \cite{kang2011automatic} first detect rain from the high/low frequency part of input images based on morphological component analysis and remove rain streaks in the high frequency layer via dictionary learning. Similarly, Huang \textit{et al.} \cite{huang2013self} and Zhu \textit{et al.} \cite{zhu2017joint} use sparse coding based methods to remove rain from a single image. Some works aim to remove rain based on low-rank representation \cite{chen2013generalized,zhang2017convolutional}. Chen \textit{et al.} \cite{chen2013generalized} generalize a low-rank model from matrix to tensor structure, which does not need the rain detection and dictionary learning stage. In addition, Li \textit{et al.} \cite{li2016rain} use a GMM trained on patches from natural images to model the background patch priors.

Recently, deep learning achieves significant success in low-level vision tasks such as image super-resolution \cite{ledig2017photo,johnson2016perceptual,niu2020single}, deblurring \cite{zhang2018adversarial,zhang2020deblurring,li2021arvo}, dehazing \cite{ren2016single,li2017aod}, which also include deraining \cite{li2019single,zhang2019image,fu2017clearing,fu2017removing,yang2017deep,zhang2018density,li2018recurrent,eigen2013restoring,qian2018attentive,zheng2019residual,zhang2021enhanced,zhang2021deep,zhang2021dual}. These methods learn a mapping between input rainy images and their corresponding clean version using CNN/RNN based models. Some of them use an attention mechanism to pay attention to depth \cite{hu2019depth}, heavy rain regions \cite{li2019heavy} or density \cite{zhang2018density}. However, to the best of our knowledge, there are few deep deraining works which try to remove rain via scene understanding \cite{long2015fully}.

\subsection{Video Deraining}

Video deraining is to obtain a clean video from an input rainy video. Compared with single image deraining, methods for video deraining can not only learn the spatial information, but also leverage temporal information in removing rain.

Traditional methods try to use prior such as the temporal context and motion information \cite{garg2004detection,garg2006photorealistic}. Researchers formulate rain streaks based on their intrinsic characteristics \cite{zhang2006rain,liu2009pixel,santhaseelan2015utilizing,brewer2008using,jiang2017novel} or propose some learning-based methods to improve the performance of deraining models \cite{chen2013rain,tripathi2012video,kim2015video,wei2017should,ren2017video}. For example, Santhaseelan \textit{et al.} \cite{santhaseelan2015utilizing} and Barnum \textit{et al.} \cite{barnum2010analysis} extract phase congruence features and Fourier domain features, respectively, to remove rain streaks. Chen \textit{et al.} \cite{chen2013rain} apply photo-metric and chromatic constraints to detect rain and utilize filters to remove rain in the pixel level.

Deep learning methods are also proposed for video deraining \cite{liu2018d3r,liu2018erase,chen2018robust,yang2019frame}. Chen \textit{et al.} \cite{chen2018robust} propose a robust deep deraining model via applying super-pixel segmentation to decompose the scene into depth consistent unites. Liu \textit{et al.} \cite{liu2018d3r} depict rain streaks via a hybrid rain model, and then present a dynamic routing residue recurrent network via integrating the hybrid model and using motion information. Yang \textit{et al.} \cite{yang2019frame} consider the additional degradation factors in the real world and propose a two-stage recurrent network for video deraining. Their model is able to capture more reliable motion information at the first stage and keep the motion consistency between frames at the second stage. Although these methods use the information of multiple rainy images, all of them extract features from a sequence of monocular frames and ignore the stereo views.

\subsection{Stereo Deraining}

Stereo images provide more information from cross views and have thus been utilized to improve the performance of various computer vision tasks, including traditional problems \cite{godard2017unsupervised,eslami2016attend,luo2016efficient} and novel tasks \cite{jeon2018enhancing,li2018depth,chen2018stereoscopic,zhou2019davanet}. However, there are few methods that leverage the stereo images to remove rain so far. Yamashit \textit{et al.} \cite{tanaka2006removal} remove the rain via utilizing disparities between stereo images to detect positions of noises and estimate true disparities of images regions hidden into rain. In order to obtain the derained left-view images, Kim \textit{et al.} \cite{kim2014stereo} warp the spatially adjacent right-view frames and subtract the warped frames from the original frames. However, these traditional methods do not consider the importance of semantic information. Meanwhile, the strong capability of learning features implied in deep neural networks is also ignored by them.

\section{The Semantic-aware Deraining Module}
\label{SADM}
The ultimate goal of our work is to recover the deraining images from their corresponding rainy versions. In order to improve the capability of our model, a semantic-aware deraining module is proposed to learn semantic features based on clean images, rainy images and semantic labels. In this section, we will first introduce the consolidation of different tasks in Sec. \ref{different_task} and how to train the proposed module based on images and semantic-annotated images in Sec. \ref{training}. Then, a semantic-rethinking loop is discussed in Sec. \ref{loop} to further enhance our module and extract powerful features.

\subsection{The Consolidation of Different Tasks}
\label{different_task}

\begin{figure}[!tb]
  \centering
\includegraphics[width=0.99\linewidth ]{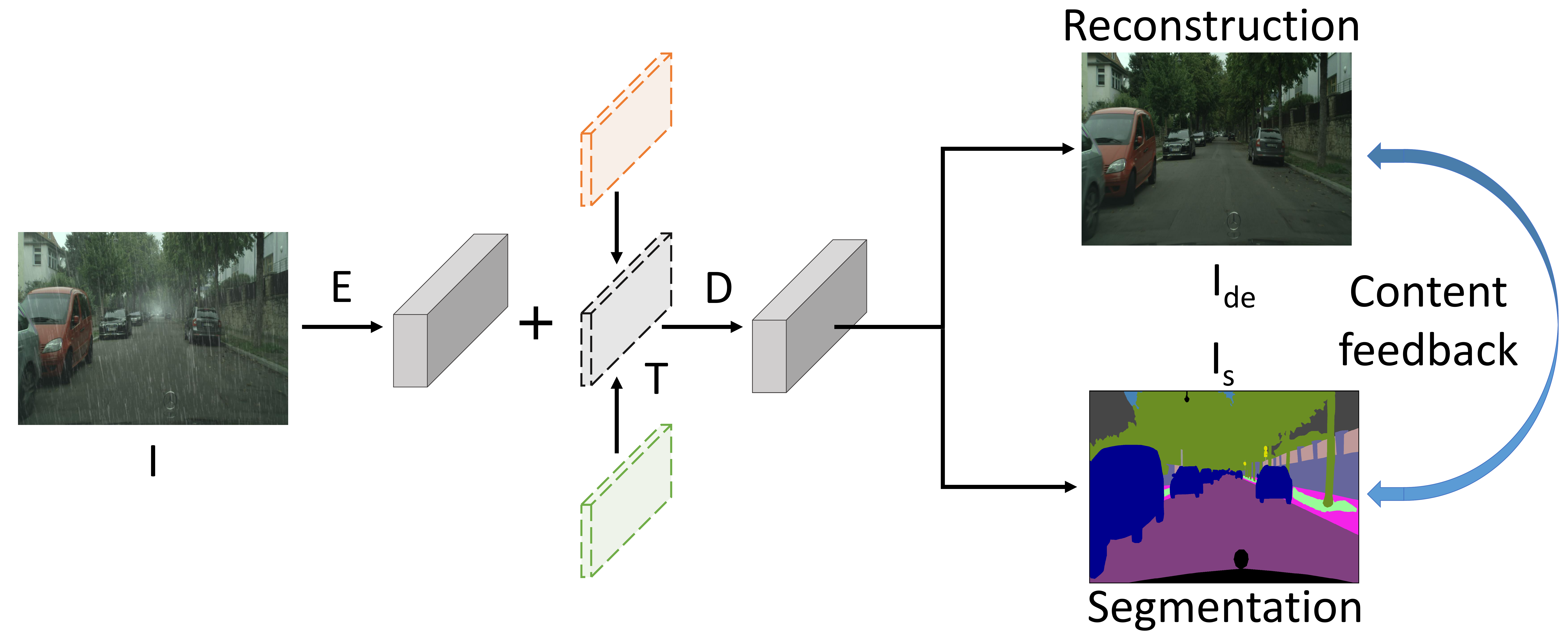}
\caption{{\bf The architecture of the proposed semantic-aware deraining module.} Rainy images are fed into the encoder $E$ to extract features. $T$ represents the task labels, which can be deraining labels or scene labels. Then the decoders $D$ generate deraining and segmentation results for different tasks.}
\label{semantic}
\end{figure}

Currently, most deep deraining methods directly learn the transformation from rainy images to the derained ones \cite{li2019single}. Inspired by \cite{hu2019depth}, which proposes a depth-aware network to jointly learn depth estimation and image deraining via two different sub-networks. In this paper, an autoencoder architecture is employed to merge different tasks in the learning stage. Fig. \ref{semantic} illustrates the architecture of the proposed module. Images are input into the encoder of the proposed module to extract semantic features $F$. Then the semantic features $F$ combined with a task label $T$ are fed into the following decoder architecture to obtain a prediction $P$ corresponding to label $T$. Based on different task labels like \textit{deraining} or \textit{scene understanding}, different outputs will be obtained. The learning stage can be formulated as
\begin{equation}
P = D(E(I), T)\,,
\end{equation}
where $E$ and $D$ are the encoder and decoder of \textit{SADM}, respectively. $I$ is the input image. $T$ represents the label of different tasks. Based on the output of the encoder and $T$, different predictions will be derived. 

The branch of image deraining can be denoted as
\begin{equation}
I_{de} = \sigma_{de} (P \ | \ T_{de})\,,
\end{equation}
where $T_{de}$ corresponds to the label of deraining image. $\sigma_{de}$ is the mapping function. 

The branch of understanding scenes can be formulated as
\begin{equation}
I_{seg} = \sigma_{seg} (P \ | \ T_{seg})\,,
\end{equation}
where $T_{seg}$ corresponds to the semantic segmentation label. $\sigma_{seg}$ is a softmax function. 

Based on the conditional architecture \cite{zhao2018dynamic}, the proposed \textit{SADM} can jointly learn scene understanding and image deraining, which can extract more powerful semantic-aware features via sharing the information learned from different tasks, therefore being beneficial to multiple tasks.

\subsection{Image Deraining and Scene Segmentation}
\label{training}

\textbf{Image Deraining.} When $T$ is set to $T_{de}$, the output of the proposed module is the deraining image. To learn the image deraining model, we compute the image reconstruction loss based on the MSE loss function:
\begin{equation}
\mathcal{L}_{de} = ||I_c - \sigma_{de} (D(E(I_{rainy}), T_{de}))||^2 \,,
\end{equation}
where $I_c$ is the clean image.

\textbf{Scene Segmentation.} Most existing deraining methods focus on pixel-level loss function and thus fail to model the geometric and semantic information. This makes it difficult for models to understand the input image and generate deraining results with favorable details. To address this problem, we remove rain from rainy images by leveraging semantic information. The learning process of scene understanding can be denoted as
\begin{equation}
\mathcal{L}_{seg} = \sigma_h(I^{gt}_{seg}, I_{seg}) \,,
\end{equation}
where $I_{seg}$ and $I^{gt}_{seg}$ indicate the scene understanding of the model and ground truth labels from auxiliary training sets. $\sigma_h$ is the cross-entropy loss function.

\begin{figure}[!tb]
  \centering
\includegraphics[width=0.99\linewidth ]{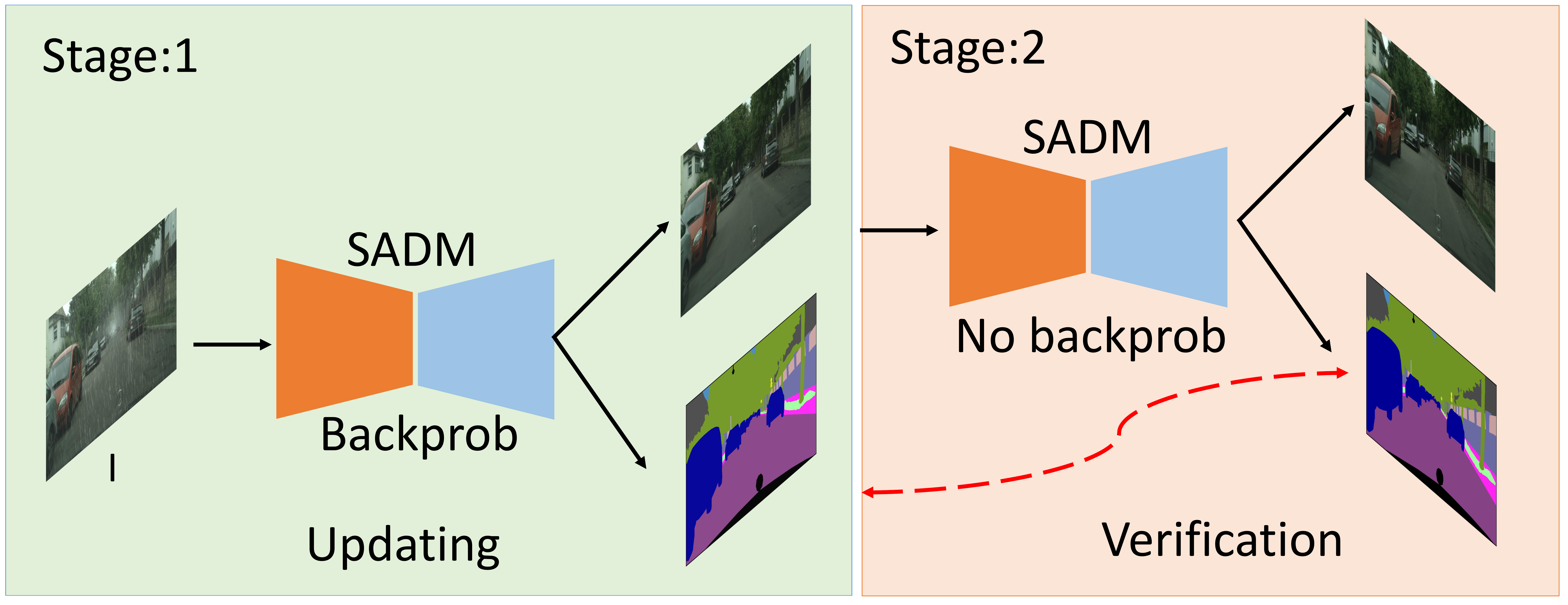}
\caption{{\bf The Semantic-rethinking Loop.} During training, rainy images are fed into \textit{SADM} to generate deraining and segmentation results in stage I. Then the deraining images are utilized to generate segmentation results again in stage II. Through comparing the two segmentation results from rainy and deraining images, \textit{SADM} can better understand scenes and remove the undesired rain. \textit{SADM}s in the  two stages share the weights.}
\label{figure_loop}
\end{figure}

\subsection{Semantic-rethinking Loop}
\label{loop}

Semantic information plays an important role in various tasks of computer vision \cite{shen2018deep,shen2020exploiting,li2020word,li2020transferring,zhang2020uc,zhang2020weakly}. In order to further enhance the semantic understanding of our model and help remove rain, a semantic-rethinking loop is proposed to refine the error-prone semantic understanding. Fig. \ref{figure_loop} illustrates its scheme. It consists of an ``updating" part and a ``verification" part, whose core architecture is the semantic-aware deraining module, which has been illustrated in Fig. \ref{semantic}.

In the training stage, the ``updating" part takes a rainy image as input, and then generates the deraining image and semantic segmentation. Loss functions introduced in above sections are calculated and then update the weights of layers in the semantic-aware deraining module. Then the deraining image obtained in the ``updating" part is fed into the ``verification" part to obtain new semantic segmentation. The semantic understanding can improve the performance of deraining, which will be demonstrated in the next section. However, rain increases the difficulty of scene understanding. Via comparing segmentation results in different parts and pushing them to be close, \textit{SADM} can better understand scenes and thus better derain. Both ``updating" and ``verification" parts employ the semantic-aware deraining module.
The main difference between the ``updating" and ``verification" parts is that the weights in semantic-aware deraining module are updated in the ``updating" part but fixed in the ``verification" part. The semantic-rethinking loop provides the content feedback from the coarse-deraining image and improves the semantic understanding of \textit{SADM}. In the testing stage, only the core semantic-aware deraining model is utilized to remove rain from images. The loss function can be noted as
\begin{equation}
\mathcal{L}_{con} = ||I^{ver}_{seg} - I^{up}_{seg}||\,,
\end{equation}
where $I^{ver}_{seg}$ and $I^{up}_{seg}$ are the semantic segmentation results from the ``verification" and ``updating" parts, respectively.

\begin{figure}[!tb]
  \centering
\includegraphics[width=0.99\linewidth ]{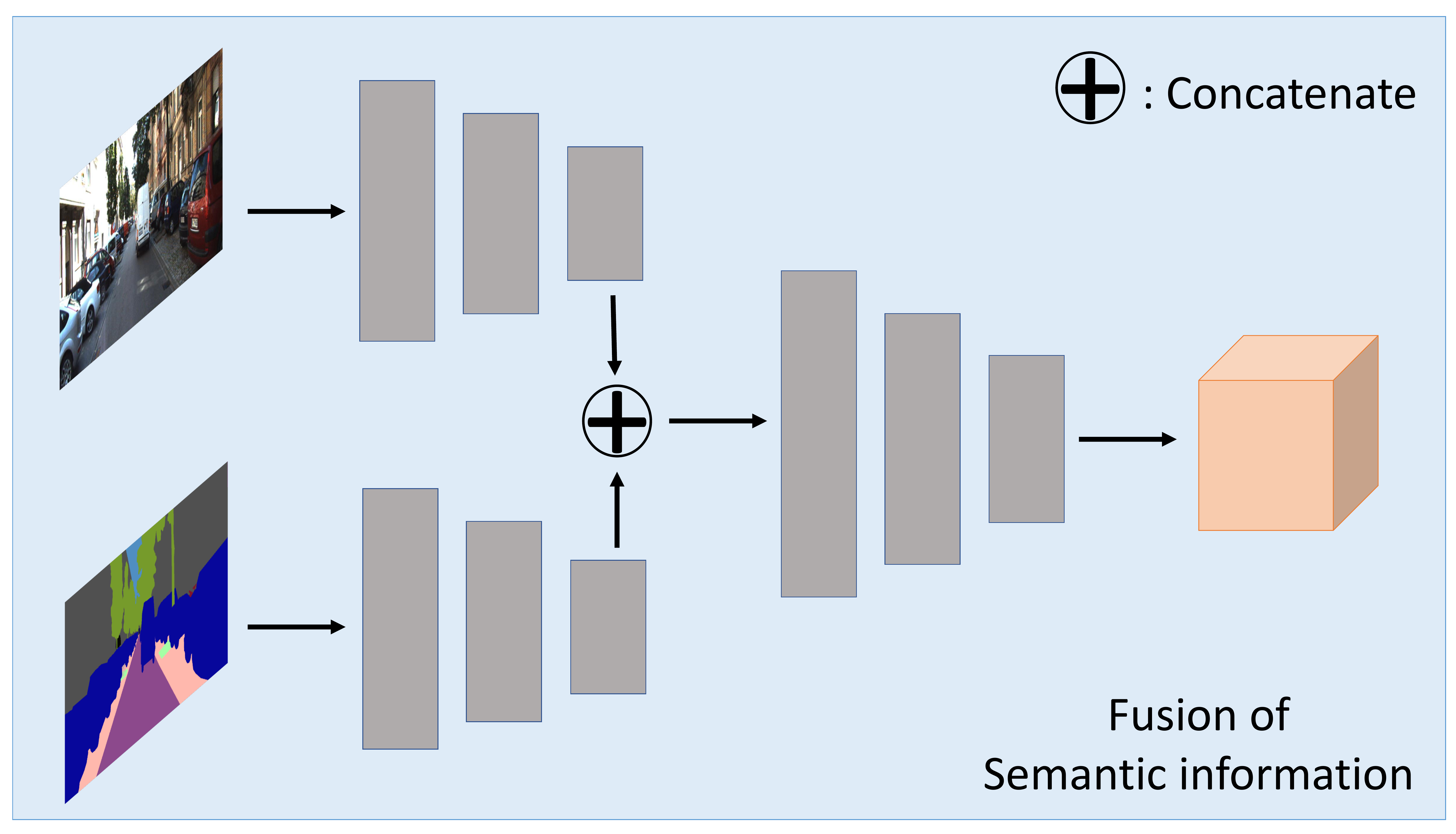}
\caption{{\bf The architecture of \textit{SFNet}.} The coarse deraining images and semantic segmentation results from \textit{SADM} are fed into \textit{SFNet} to generate features volume with semantic information.}
\label{figure_semantic_fusion}
\end{figure}

\section{The Paired Rain Removal Network}

In order to remove rain from stereo images, we further present a \textit{PRRNet} based on \textit{SADM}. The overall of the proposed network will firstly be introduced in Sec. \ref{overall}, and then two core sub-networks will be discussed in Sec. \ref{Semantic-FusionNet}  and \ref{View-FusionNet}. Finally, the objective functions to train the proposed model will be presented in Sec. \ref{object}. Sec. \ref{prrnet:implementation} provides implementation details.

\subsection{Network Architecture}
\label{overall}

\textit{PRRNet} consists of three sub-networks, \textit{i.e.}, \textit{SADM}, Semantic-Fusion Net (\textit{SFNet}) and View-Fusion Net (\textit{VFNet}). \textit{SADM} is introduced in Sec. \ref{SADM} to jointly remove rain and understand semantic information. Semantic-Fusion Net is utilized to combine the semantic information with coarse deraining images, while View-Fusion Net is to combine information from different views to obtain final deraining images. Due to the above-mentioned stereo semantic-aware deraining module, the proposed \textit{PRRNet} simultaneously considers cross views and semantic information to help remove rain from images.

\subsection{SFNet}
\label{Semantic-FusionNet}

The architecture of \textit{SFNet} is shown in Fig. \ref{figure_semantic_fusion}. The input is semantic segmentation and coarse deraining images from \textit{SADM}. Given that the semantic information can help remove rain, we first process them individually and concatenate them, and then forward them into the following layers, to generate feature volume, which is utilized for generating final deraining results.

\begin{figure}[!tb]
  \centering
\includegraphics[width=0.99\linewidth ]{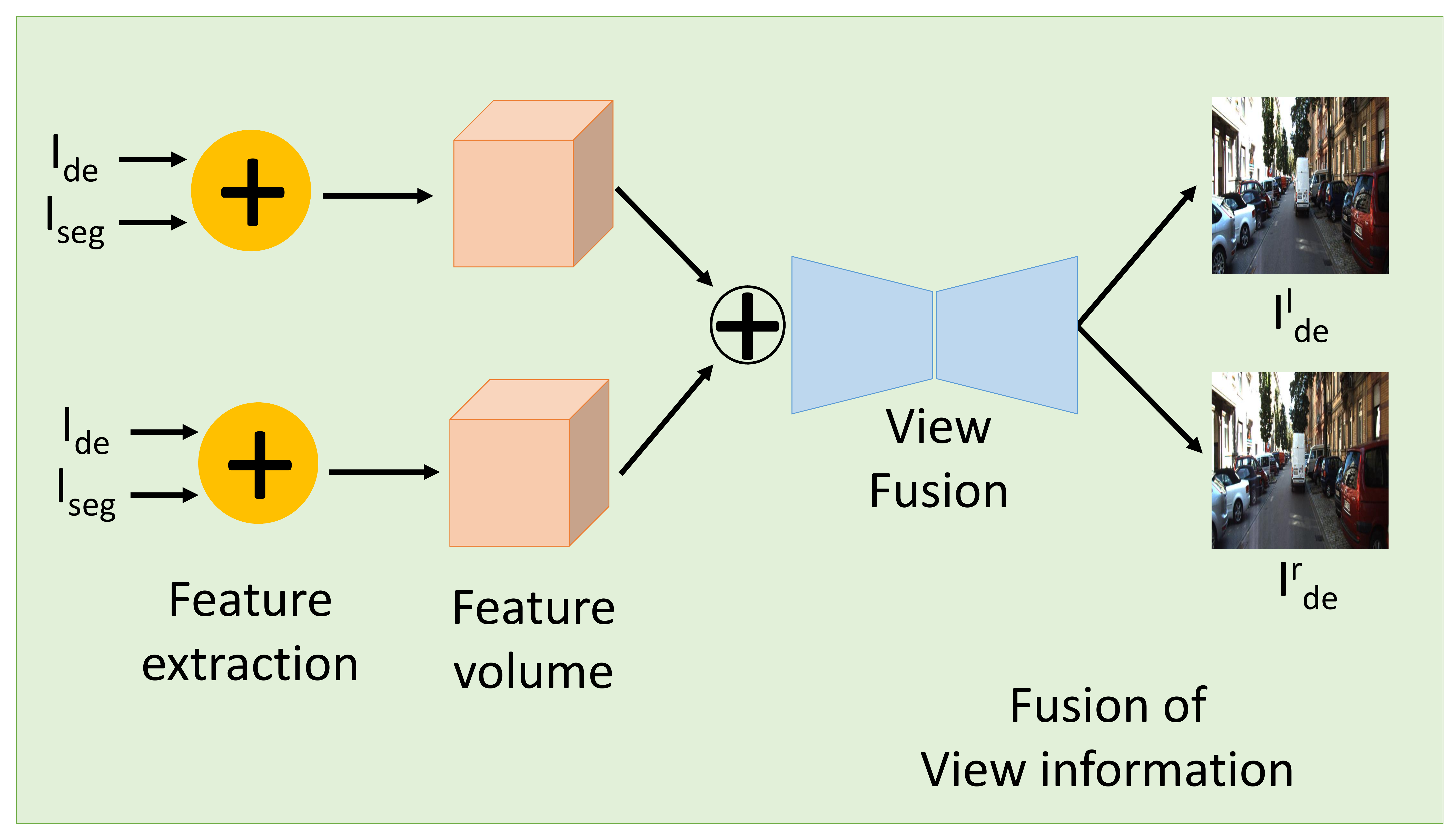}
\caption{{\bf The architecture of \textit{VFNet}.} Features volumes from stereo images are fused to generate final stereo deraining images.}
\label{figure_view_fusion}
\end{figure}

\subsection{VFNet}
\label{View-FusionNet}
Fig. \ref{figure_view_fusion} illustrates the architecture of \textit{VFNet}. The input is extracted fusion features from \textit{SFNet}. The features extracted from the right view are helpful to remove the rain in the left-view image. Similarly, removing the rain from the right-view image also takes advantage of features captured from the left-view image. Through the \textit{VFNet}, the final finer deraining stereo images are obtained. The loss function in this part can be denoted as
\begin{equation}
\mathcal{L}_{view} = ||I^{left}_{de} - I^{left}_{gt}|| + ||I^{right}_{de} - I^{right}_{gt}||,
\end{equation}
where $I^{left}_{de}$ and $I^{right}_{de}$ are stereo deraining images from \textit{VFNet}, respectively. $I^{left}_{gt}$ and $I^{right}_{gt}$ are the clean version of the stereo images.

\subsection{Objective Functions}
\label{object}
The loss function consists of two kinds of data terms, which are calculated based on semantic understanding and deraining reconstruction images. The final loss function can be written as
\begin{equation}
\mathcal{L}_{f} = \mathcal{L}_{de} + \lambda_1 \mathcal{L}_{seg} + \lambda_2 \mathcal{L}_{con} + \lambda_3 \mathcal{L}_{view} ,
\end{equation}
where $\mathcal{L}_{de}$ and $\mathcal{L}_{view}$ are utilized to remove the rain from rainy images, and $\mathcal{L}_{seg}$ and $\mathcal{L}_{con}$ push the model to understand scenes better, which are helpful for stereo deraining. $\lambda_1$, $\lambda_2$ and $\lambda_3$ are three parameters to balance different loss functions, which are set as $1.0$, $0.2$ and $1.0$, respectively.

\subsection{Implementation Details}
\label{prrnet:implementation}
\textit{SADM} is an encoder-decoder architecture. The encoder network consists of $13$ CNN layers, which is initialized by a VGG16 network pre-trained for object classification. The decoder also has $13$ CNN layers. \textit{SFNet} contains three CNN layers ($32 \times 3 \times 3$) which are utilized to fuse the semantic information. \textit{VFNet} contains five ResBlocks \cite{he2016deep} to generate final deraining results. Each ResBlock consists of three CNN layers of $64 \times 3 \times 3$ kernels and two ReLU activation layers. The proposed \textit{PRRNet} is trained with Pytorch library. The base learning rate is set to $10^{-4}$ and then declined to $10^{-5}$. The model is updated with the batch size of 2 during the training stage.

\begin{figure*}[!tb]
  \centering
  \subfigure[]{
    \label{}
    \includegraphics[width=0.8\linewidth ]{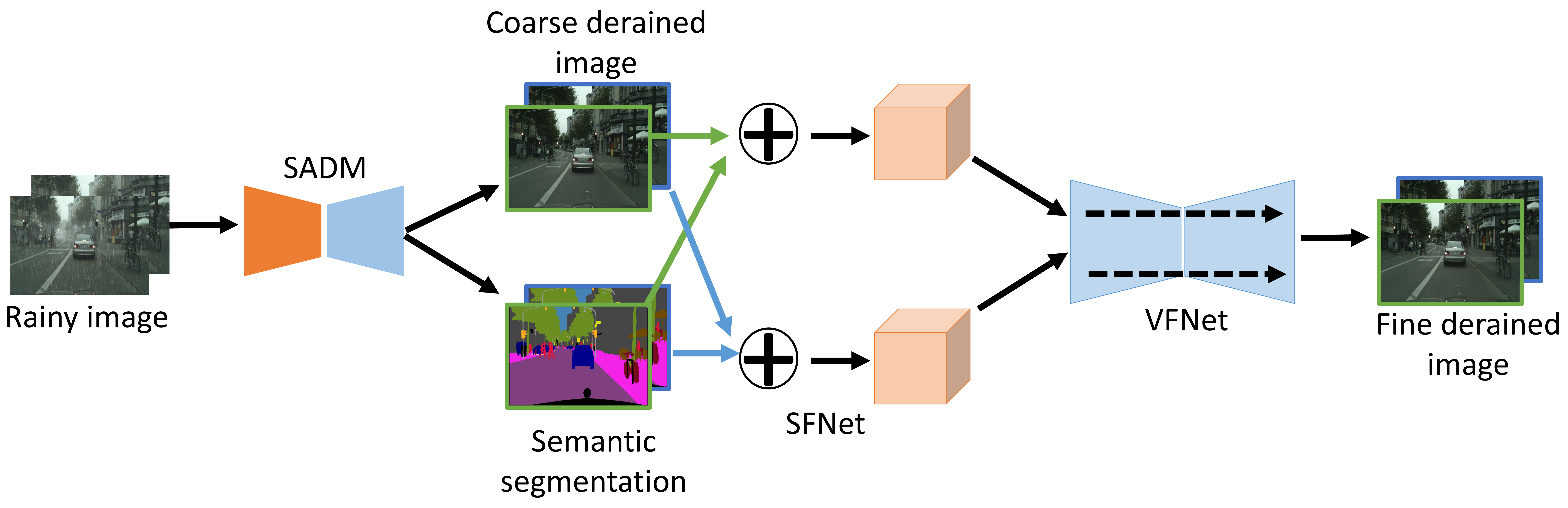}}
    \subfigure[]{
    \label{}
    \includegraphics[width=0.8\linewidth ]{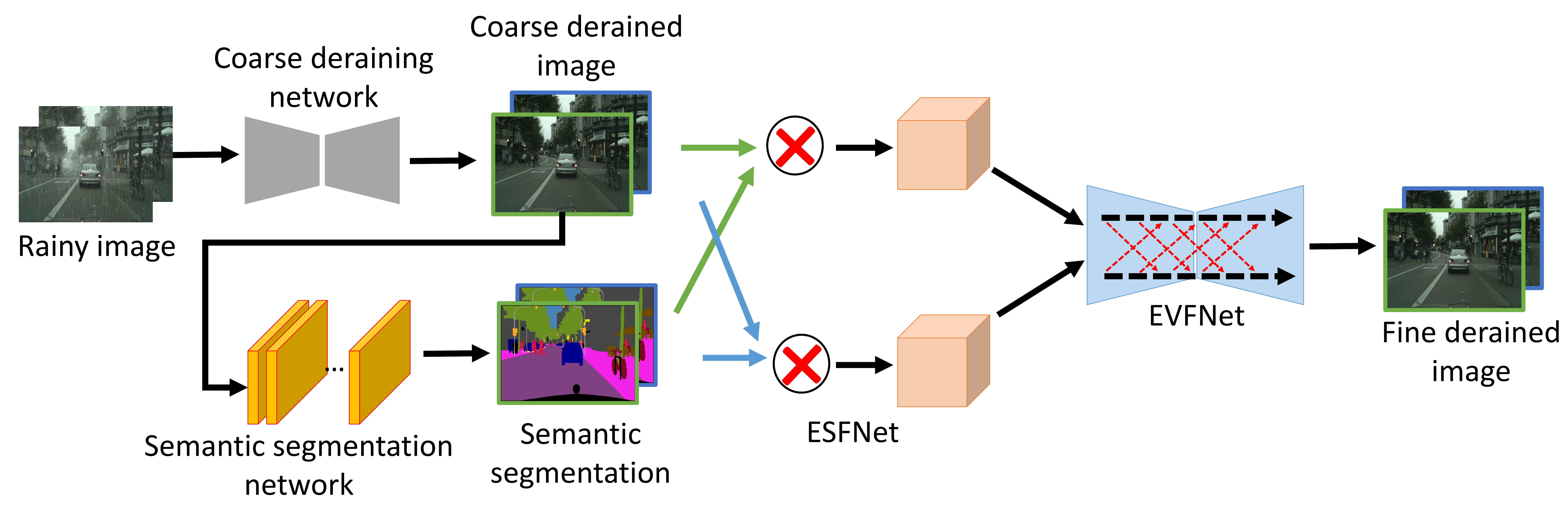}}
\caption{ {\bf Overview of the proposed model.} The top figure is the framework of the state-of-the-art stereo dearining method \cite{zhang2020beyond}, which directly predict the semantic labels from rainy images to help their following process. The bottom figure is our proposed enhanced version, which first reduces the rain streaks via a coarse deraining network. In addition, the proposed network replaces the \textit{SFNet} and \textit{VFNet} with \textit{ESFNet} and \textit{EVFNet} in a semantic-attentional and parallel cooperation ways, respectively.}
  \label{figure_new_overview}
\end{figure*}

\section{The Enhanced Paired Rain Removal Network}

Currently, there exist several well pre-trained networks for semantic segmentation which can extract satisfactory semantic information. However, the unwanted rain streaks make the labels extracted from the rainy images incorrect. In this section, we introduce an Enhanced Paired Rain Removal Network (EPRRNet) under the ``coarse-to-fine" scheme to reduce the effect of rain streaks and make better use of the semantic prior. We first give an overview of the proposed \textit{EPRRNet} model. Then, we introduce four core modules, coarse deraining network, semantic segmentation network, Enhanced SFNet (\textit{ESFNet}) and Enhanced VFNet (\textit{EVFNet}).

\subsection{Overall}

In our preliminary work \cite{zhang2020beyond}, we build a \textit{SADM} to extract semantic labels from the input rainy images and then apply an \textit{SFNet} and a \textit{VFNet} to fuse semantic prior and information from stereo views, respectively. However, the unwanted rain streaks make the labels extracted from the rainy images incorrect. In this section, we improve from the following aspects:

\begin{itemize}

\item
To reduce the effect of rain streaks and extract better semantic labels, we first use a coarse deraining network to obtain coarse derained results, and then apply a pre-trained network to predict the semantic labels.

\item
The \textit{SFNet} is replaced by an \textit{ESFNet} to fuse the semantic prior. Specially, we apply a semantic-guided attention mechanism to learn semantic-attentional features, and then combine with non-semantic-attention features to remove rain streaks.

\item
The \textit{VFNet} is replaced by an \textit{EVFNet} to fuse the information from stereo views. The proposed \textit{EVFNet} is a parallel stereo network. Stereo images are fed into networks to extract two different types of information. Different from the \textit{VFNet} which fuses the high-level features from two views to obtain the final derained images, the cooperation between stereo images in our proposed \textit{EVFNet} is carried out from the low-level layers to high-level layers in a parallel way.

\end{itemize}

The differences between the \textit{PRRNet} and \textit{EPRRNet} are shown in the Fig. \ref{figure_new_overview}.

\begin{figure*}[!tb]
  \centering
\includegraphics[width=0.8\linewidth ]{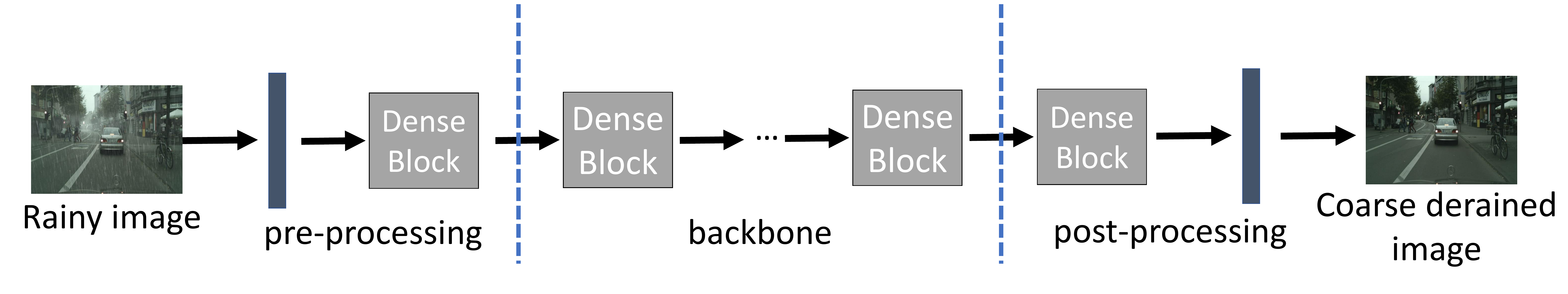}
\caption{{\bf The architecture of the coarse deraining network.} It consists of three modules, \textit{i.e.}, pre-processing module, backbone module and post-processing module. Its input and output are the rainy images and coarse derained results, respectively.}
\label{figure_coarse_derain}
\end{figure*}

\subsection{Coarse Deraining Network}
To reduce the negative effect of the rain streaks, we build a coarse deraining network to obtain the coarse deraining results:
\begin{equation}
I^{c}_{de} = G (I_{i}) \,,
\end{equation}
where $I_{i}$ and $G$ are the input rainy images and the coarse deraining network, respectively. 

The coarse deraining network is similar to the \textit{SADM} with several variations. Firstly, as the output is only coarse deraining images, we use a one-branch output network to replace the two-branch model. Secondly, we use DenseBlocks to build the coarse deraining network instead of AutoEncoder. Specially, it consists of three modules, \textit{i.e.}, the pre-processing module, the backbone module and the post-processing module. The overall architecture of the proposed network is shown in the Fig. \ref{figure_coarse_derain}. The pre-processing modules includes a CNN and a DenseBlock to generate 16 feature maps. The backbone module consists of five DenseBlocks, whose input is the output of the pre-processing model. The details of DenseBlock is similar to \cite{zhang2020deblurring}. The feature maps in the backbone module is also set to $16$. We use ReLU as the activation function for all convolutional layers. In order to restore derained images with better details, we build a post-processing module to remove the artifacts and improve the quality of the restored images.  It consists of a DenseBlock and a convolutional layer.

\begin{figure*}[!tb]
  \centering
\includegraphics[width=0.8\linewidth ]{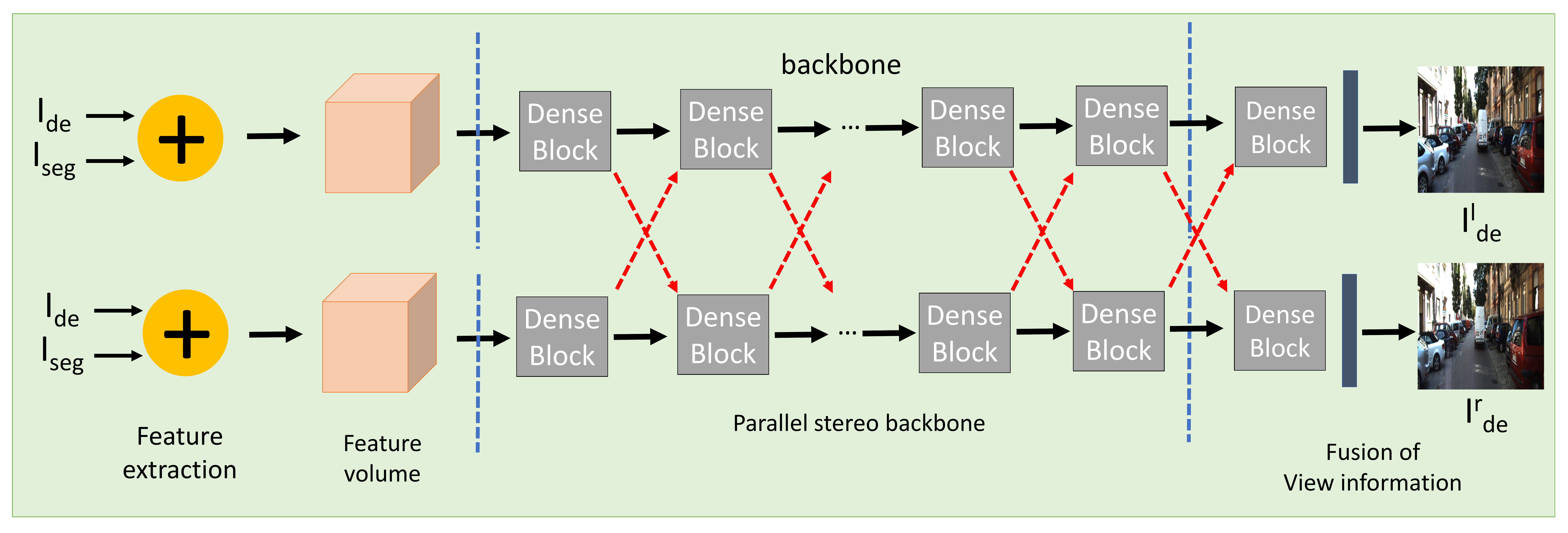}
\caption{{\bf The architecture of the EVFNet.} The input is the feature volume extracted from stereo images. Then both of them are fed into a parallel stereo backbone. The output of each DenseBlock from the left image is concatenated with the output from the right image. In this way, the communication between stereo images are conducted from low-level to high-level layers.}
\label{figure_evfnet}
\end{figure*}

\begin{figure}[!tb]
  \centering
\includegraphics[width=0.99\linewidth ]{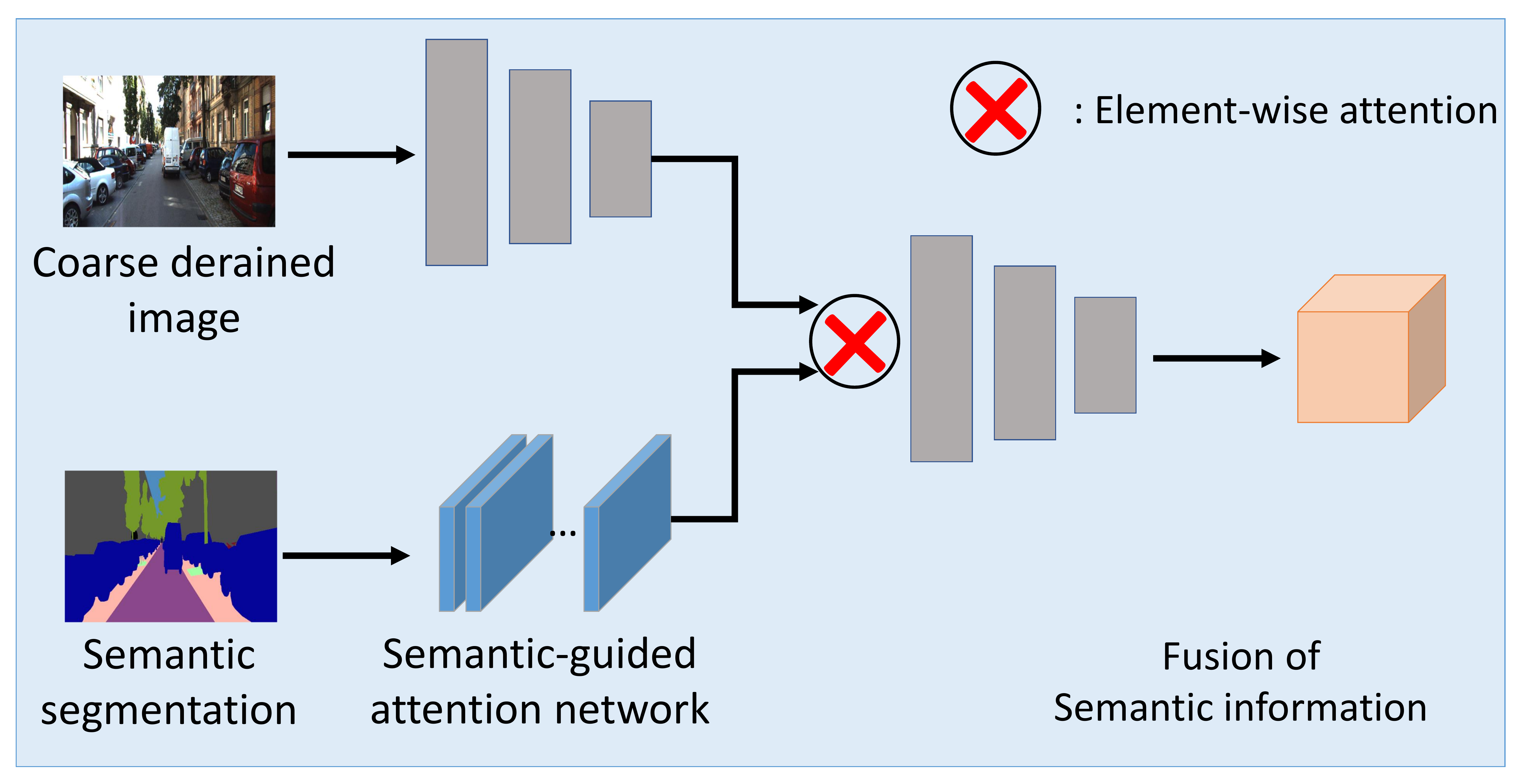}
\caption{{\bf The architecture of the ESFNet.} The input is coarse derained images and semantic labels. The semantic-attentional features are calculated and fed into the following layers to help remove rain streaks.}
\label{figure_esfnet}
\end{figure}

\subsection{Semantic Segmentation Network}

We use a hierarchical multi-scale model \cite{tao2020hierarchical} to predict the semantic information. The output of the coarse deraining network is fed into the semantic segmentation network $S$ to generate the predication of semantic labels $I_{seg}$:
\begin{equation}
I_{seg} = S (I^{c}_{de}) \,,
\end{equation}
where $I_{seg} \in \mathbb{R}^{H \times W \times K}$. $H$, $W$ and $K$ represent the height, weight of input images, and the number of semantic classes. The prediction of semantic information encode different objects of input images like face, cars and buildings and can be applied as priors for recovering derained images. In this paper, the $K$ is set as 30 to obtain 30 semantic labels for fine deraining process.

\subsection{Enhanced SFNet}

As the semantic information is able to provide strong global prior for images restoration, we take the estimated semantic labels as a guidance to learn semantic-attentional features, which are concatenated with features without semantic attention, to help remove rain streaks.

Specially, we first use a pre-processing module to encode the input coarse derained images, which are then fed into a CNN network to obtain attention weights via modeling the semantic-guided attention mechanism. Similar to \cite{hu2019depth}, the CNN network contains 7 convolutional layers. Its output is a set of attention weights $\{A_i, A_2, ..., A_n\}$, which are fed into a Softmax layer to normalize and obtain the attention weights $\{
W_i, W_2, ..., W_n\}$, which is formulated as:
\begin{equation}
W_i =  \frac{e^{A_c}}{\sum_{c=1}^{n} e^{A_c}} \,,
\end{equation}
where $c$ is the channel of features. Fig. \ref{figure_esfnet} shows the process of fusing semantic-attentional features. 

\subsection{Enhanced VFNet}

In order to make better use of the information from stereo views, we introduce a parallel stereo network to convolve features from stereo images. Different from the above VFNet which combines stereo information in the last layers, the \textit{EVFNet} makes use of stereo information from low-level to high-level layers. 

Fig. \ref{figure_evfnet} illustrates the overall architecture of EVFNet, the backbone is similar to the coarse deraining network. The EVFNet contains two share-weights streams to process input features in a parallel way. It divides the input two stereo images into two sub-nets and use a regular convolution over each sub-net to extract features.
In order to exchange the information across stereo images, a stereo-image fusion module is applied between two streams as Fig. \ref{figure_evfnet}. Specially, the output of each DenseBlock from the top sub-net is combined with the output of each DenseBlock from the bottom sub-net. Finally, the output of backbone is fed into the post-processing modules to obtain final derained results. 
%

\section{Experiments}

\subsection{Datasets}
\textbf{RainKITTI2012 dataset.} To the best of our knowledge, there are no benchmark datasets that provide stereo rainy images and their corresponding ground-truth clean versions. In this paper, we first use Photoshop to create a synthetic RainKITTI2012 dataset based on the public KITTI stereo 2012 dataset \cite{geiger2013vision}. The training set contains $4,062$ image pairs from various scenarios, and the testing set contains $4,085$ image pairs. The size of images is $1242 \times 375$.

\textbf{RainKITTI2015 dataset.} The KITTI2015 dataset is another set from the KITTI stereo 2015 dataset \cite{geiger2013vision}. Therefore, we also synthesize a RainKITTI2015 dataset, whose training set and testing set contain $4,200$ and $4,189$ pairs of images, respectively.

\textbf{Cityscapes dataset.} Cityscapes dataset is utilized as the semantic segmentation data to train \textit{PRRNet}. This dataset contains various urban street scenes and provides images with pixel-wise segmentation labels. It includes $2,975$ images and their corresponding ground truth semantic labels.

\textbf{RainCityscapes dataset.} This dataset is built by Hu \textit{et al.} \cite{hu2019depth} based on the Cityscapes dataset \cite{cordts2016cityscapes}. The training set contains $9,432$ rainy images and the corresponding clean images and depth labels. For evaluation, the testing set contains $1,188$ images. We use this dataset to evaluate the performance of monocular deraining.


\subsection{Ablation Study}
\begin{figure}[!tb]
  \centering
  \subfigure[Input]{
    \label{ablation:a}
    \includegraphics[width=0.45\linewidth ]{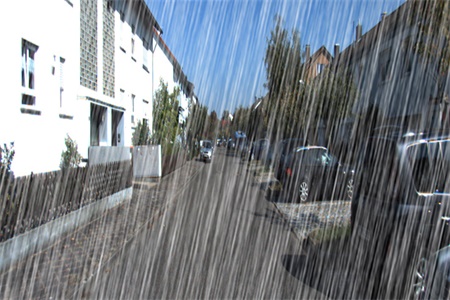}}
    \subfigure[PRRNet(D)]{
    \label{ablation:b}
    \includegraphics[width=0.45\linewidth ]{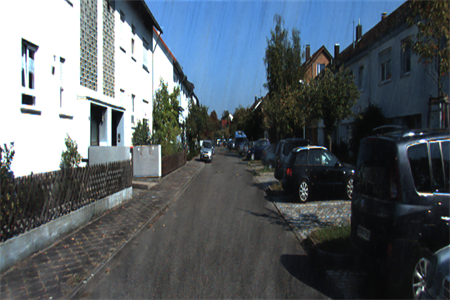}}
    \subfigure[PRRNet(D+S)]{
    \label{ablation:c}
    \includegraphics[width=0.45\linewidth ]{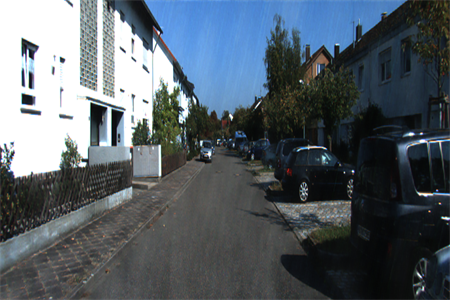}}
    \subfigure[PRRNet(D+S+L)]{
    \label{ablation:d}
    \includegraphics[width=0.45\linewidth ]{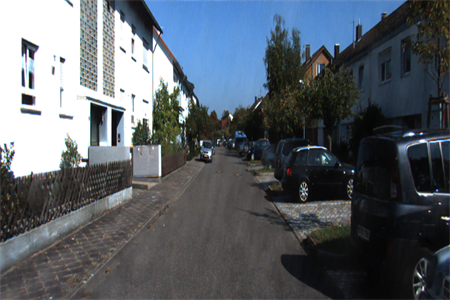}}
    \subfigure[EPRRNet(monocular)]{
    \label{ablation:d}
    \includegraphics[width=0.45\linewidth ]{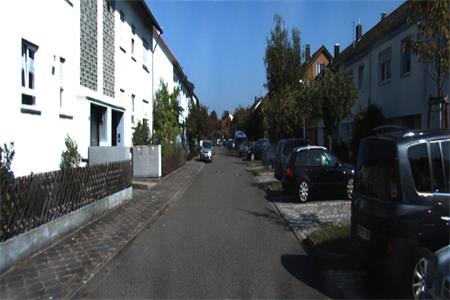}}
    \subfigure[PRRNet(stereo)]{
    \label{ablation:e}
    \includegraphics[width=0.45\linewidth ]{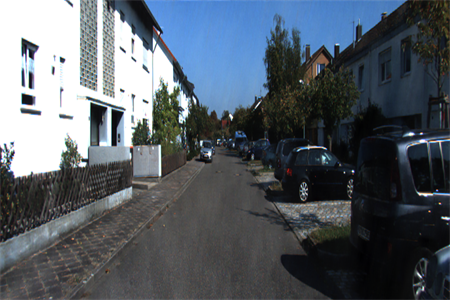}}
    \subfigure[EPRRNet(stereo)]{
    \label{ablation:e}
    \includegraphics[width=0.45\linewidth ]{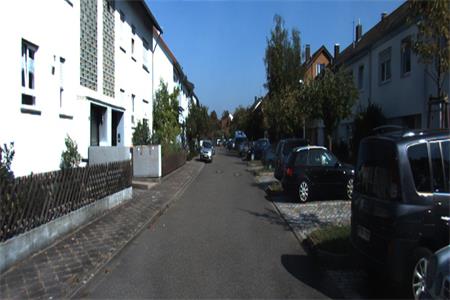}}
  \subfigure[Ground truth]{
    \label{ablation:f}
    \includegraphics[width=0.45\linewidth ]{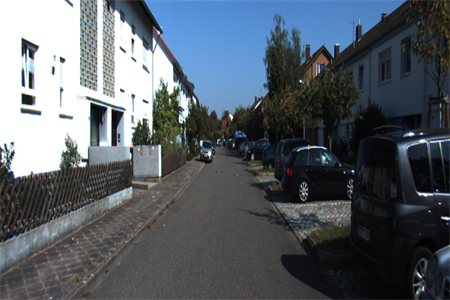}}
\caption{ \bf Exemplar results of deraining evaluation of different variant models on the dataset of RainKITTI2012.}
  \label{figure_ablation}
\end{figure}

The proposed \textit{PRRNet} takes advantage of semantic information to remove rain from images. In order to show the effectiveness of semantic information, we compare the performance of our model with the one which is trained without semantic information. Another advantage of \textit{PRRNet} is that it fuses the varying information in corresponding pixels across two stereo views to remove rain. Therefore, we also compare models trained on monocular and stereo images. 
Table \ref{table_ablation} reports the quantitative comparison results on the dataset of RainKITTI2012. Fig. \ref{figure_ablation} shows the exemplar qualitative comparison results on the same dataset. \textit{PRRNet(D)} is the model trained on monocular images with the single deraining task. \textit{PRRNet(D+S)} is the one trained on monocular images with both deraining and segmentation tasks. \textit{PRRNet(D+S+L)} is the model trained on monocular images with the above two tasks plus the semantic-rethinking loop. \textit{PRRNet(stereo)} is our full model trained based on stereo images. 
Moreover, in order to make better use of semantic information and stereo views, we also build two enhanced versions of \textit{PRRNet}, \textit{i.e.}, \textit{EPPRNet(monocular)} and \textit{EPPRNet(stereo)}, whose performance results are also shown in Table \ref{table_ablation} and Fig. \ref{figure_ablation}.

\begin{table}[!tb]
  \centering 
    \caption{\it Ablation study results on the RainKITTI2012 dataset. Both PSNR and SSIM values are reported.}
    \setlength\tabcolsep{5.0pt}
    \begin{tabular}{l |  c c }
    \toprule
    Methods &  PSNR & SSIM \\
    \hline
    \textit{PRRNet(D)}  & 30.71 & 0.923 \\
    \textit{PRRNet(D+S)}  & 31.56 & 0.928 \\
    \textit{PRRNet(D+S+L)}  & 31.89 & 0.930  \\
    \textit{EPRRNet(monocular)}  & 32.38 & 0.935 \\
    \hline
    \textit{PRRNet(stereo)} & 33.01 & 0.936 \\
    \textit{EPRRNet(stereo)}  & 34.13 & 0.947 \\
    \bottomrule
    \end{tabular}%
    \label{table_ablation}
\end{table}%

\begin{figure*}[tb]
  \centering
  \subfigure[Left input]{
    \label{kitti2012:a}
    \includegraphics[width=0.18\linewidth ]{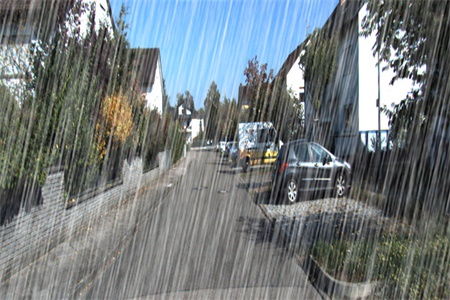}}
    \subfigure[DAF-Net]{
    \label{kitti2012:b}
    \includegraphics[width=0.18\linewidth ]{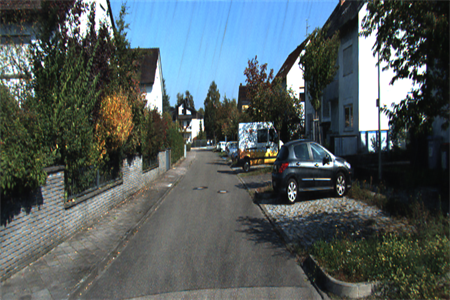}}
    \subfigure[DeHRain]{
    \label{kitti2012:c}
    \includegraphics[width=0.18\linewidth ]{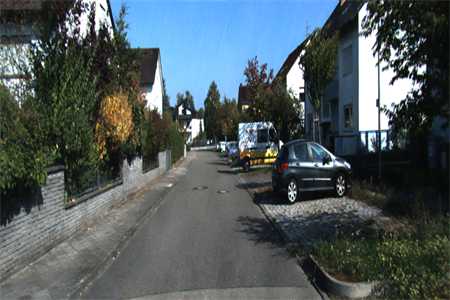}}
    \subfigure[Ours]{
    \label{kitti2012:d}
    \includegraphics[width=0.18\linewidth ]{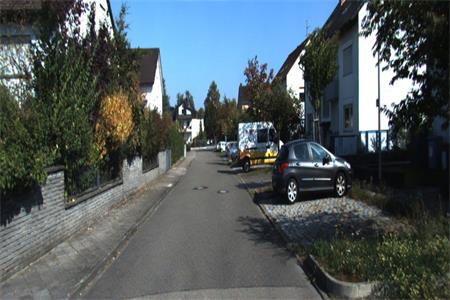}}
    \subfigure[Ground truth]{
    \label{kitti2012:d2}
    \includegraphics[width=0.18\linewidth ]{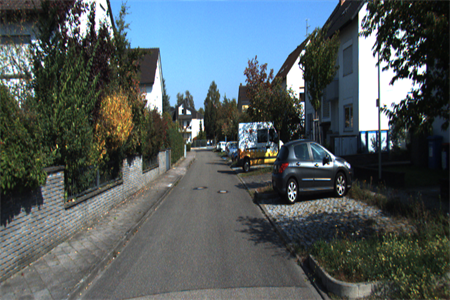}}
    \subfigure[Right input]{
    \label{kitti2012:e}
    \includegraphics[width=0.18\linewidth ]{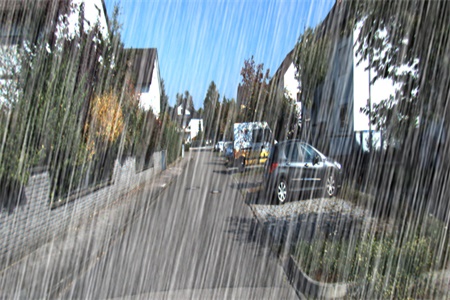}}
  \subfigure[DAF-Net]{
    \label{kitti2012:f}
    \includegraphics[width=0.18\linewidth ]{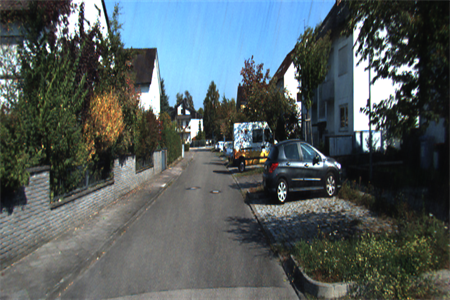}}
    \subfigure[DeHRain]{
    \label{kitti2012:g}
    \includegraphics[width=0.18\linewidth ]{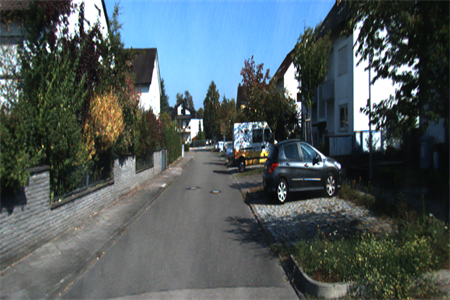}}
    \subfigure[Ours]{
    \label{kitti2012:h}
    \includegraphics[width=0.18\linewidth ]{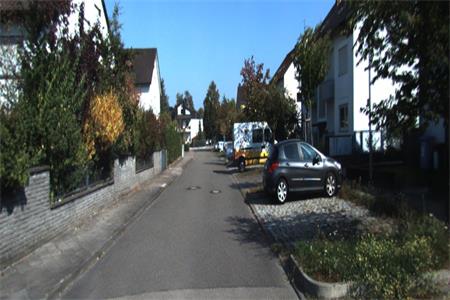}}
    \subfigure[Ground truth]{
    \label{kitti2012:h2}
    \includegraphics[width=0.18\linewidth ]{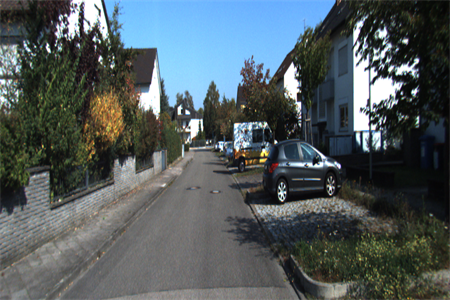}}
\caption{ \bf Exemplar results of qualitative evaluation of current SOTA models on RainKITTI2012.}
  \label{figure_kitti2012}
\end{figure*}

\begin{figure*}[tb]
  \centering
  \subfigure[Left input]{
    \label{kitti2015:a}
    \includegraphics[width=0.18\linewidth ]{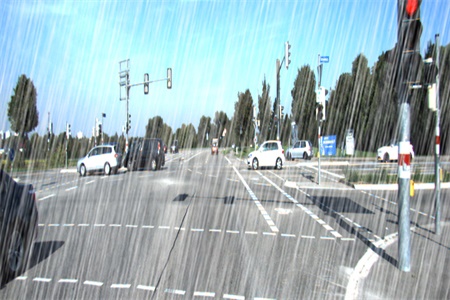}}
    \subfigure[DAF-Net]{
    \label{kitti2015:b}
    \includegraphics[width=0.18\linewidth ]{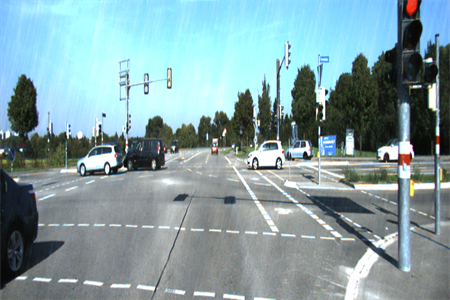}}
    \subfigure[DeHRain]{
    \label{kitti2015:c}
    \includegraphics[width=0.18\linewidth ]{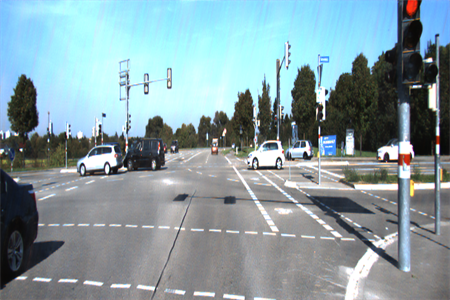}}
    \subfigure[Ours]{
    \label{kitti2015:d}
    \includegraphics[width=0.18\linewidth ]{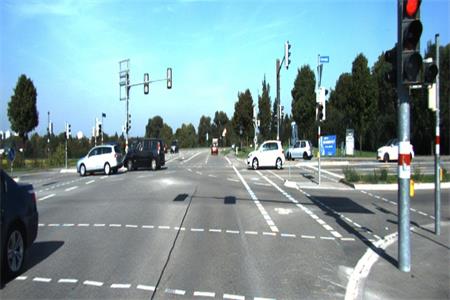}}
    \subfigure[Ground truth]{
    \label{kitti2015:d2}
    \includegraphics[width=0.18\linewidth ]{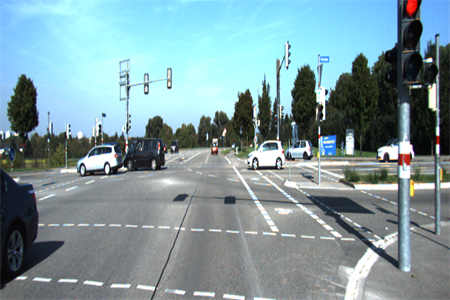}}
    \subfigure[Right input]{
    \label{kitti2015:e}
    \includegraphics[width=0.18\linewidth ]{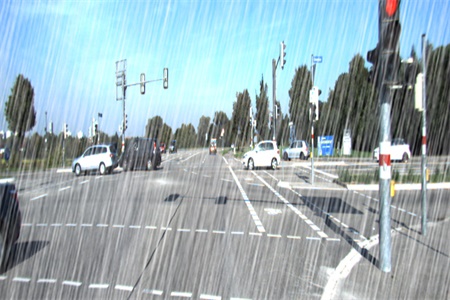}}
  \subfigure[DAF-Net]{
    \label{kitti2015:f}
    \includegraphics[width=0.18\linewidth ]{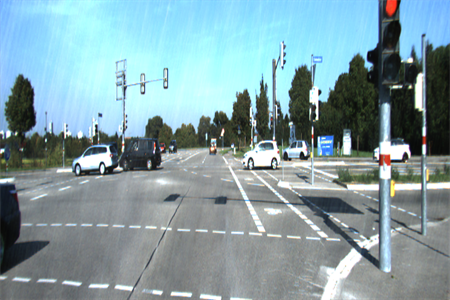}}
    \subfigure[DeHRain]{
    \label{kitti2015:g}
    \includegraphics[width=0.18\linewidth ]{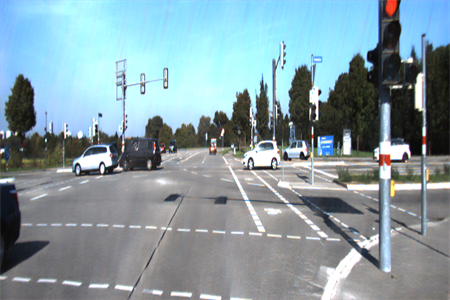}}
    \subfigure[Ours]{
    \label{kitti2015:h}
    \includegraphics[width=0.18\linewidth ]{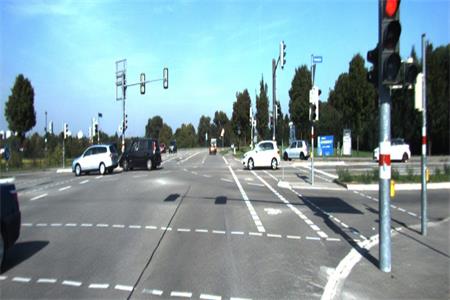}}
    \subfigure[Ground truth]{
    \label{kitti2015:h2}
    \includegraphics[width=0.18\linewidth ]{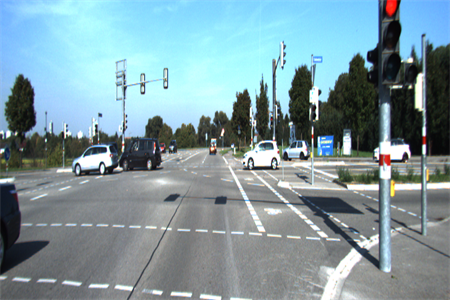}}
\caption{ \bf Exemplar results of qualitative evaluation of current SOTA models on RainKITTI2015.}
  \label{figure_kitti2015}
\end{figure*}

The results in Table \ref{table_ablation} suggest that, the plain \textit{PRRNet(D)} accomplishes the task fairly well. Additionally considering the semantic segmentation task, \textit{PRRNet(D+S)} improves the performance. With the semantic-rethinking loop, the results are further improved by \textit{PRRNet(D+S+L)}. However, the improvement is not as significant as that from \textit{PRRNet(D+S+L)} to \textit{PRRNet(stereo)} in the stereo case. This is also verified by the qualitative results in Fig. \ref{figure_ablation}. Additional components incrementally improve the visibility of the input image, and the image generated by \textit{PRRNet(stereo)} is the closest to the ground truth. The proposed \textit{EPRRNet} achieves better performance. 
Specially, in the monocular case, \textit{EPRRNet(monocular)} achieves the best perfromance. But it is still inferior to the \textit{PRRNet(stereo)}. We suspect that the cross-view information indeed provides more cues for deraining. The \textit{EPRRNet(stereo)} is better than \textit{PRRNet(stereo)}, and achieves the best performance among all the variants. This demonstrates the advantage of using semantic prior and parallel stereo network.

\subsection{Stereo Deraining}
We quantitatively and qualitatively compare our \textit{PRRNet} with current state-of-the-art methods, which include DDN \cite{yang2017deep}, DID-MDN \cite{zhang2018density}, DAF-Net \cite{hu2019depth} and DeHRain \cite{li2019heavy}. Table \ref{table_kitti2012} and Table \ref{table_kitti2015} show the quantitative results on our synthesized RainKITTI2012 and RainKITTI2015 datasets, respectively. In both tables, our monocular versions, \textit{PRRNet(monocular)} and \textit{EPRRNet(monocular)}, outperform the existing state-of-the-art methods, with remarkable gain. The model \textit{PRRNet(stereo)} and \textit{EPRRNet(stereo)} achieve the best performance with additional improvement. This demonstrates the superiority of stereo deraining over monocular deraining.

Figs. \ref{figure_kitti2012} and \ref{figure_kitti2015} compare the qualitative performance of our method \textit{EPRRNet(stereo)} and various state-of-the-art methods. The results produced by our method exhibit the smallest portion of artifacts, by referring to the ground truth.

\begin{table}[tb]
  \centering 
    \caption{\it Quantitative evaluation on the RainKITTI2012 dataset.}
    \setlength\tabcolsep{5.0pt}
    \begin{tabular}{l |  c c }
    \toprule
    Methods &  PSNR & SSIM \\
    \hline
    DDN \cite{fu2017removing} & 29.43 & 0.904  \\
    DID-MDN \cite{zhang2018density} & 29.14 & 0.901 \\
    DAF-Net \cite{hu2019depth} & 30.44 & 0.914 \\
    DeHRain \cite{li2019heavy} & 31.02 & 0.923 \\
    \hline
    \textit{PRRNet(monocular)} & 31.89 & 0.930  \\
    \textbf{\textit{EPRRNet (monocular)}}  & \textbf{32.38} & \textbf{0.935} \\
    \textit{PRRNet(stereo)} & 33.01 & 0.936  \\
    \textbf{\textit{EPRRNet (stereo)}}  & \textbf{34.13} & \textbf{0.947} \\
    \bottomrule
    \end{tabular}%
    \label{table_kitti2012}
\end{table}%

\begin{table}[tb]
  \centering 
    \caption{\it Quantitative evaluation on the RainKITTI2015 dataset.}
    \setlength\tabcolsep{5.0pt}
    \begin{tabular}{l |  c c }
    \toprule
    Methods &  PSNR & SSIM \\
    \hline
    DDN \cite{fu2017removing} & 29.23 & 0.906 \\
    DID-MDN \cite{zhang2018density} & 28.97 & 0.899 \\
    DAF-Net \cite{hu2019depth} & 30.17 & 0.915 \\
    DeHRain \cite{li2019heavy} & 30.84 & 0.921 \\
    \hline
    \textit{EPRRNet(monocular)} & 31.64 & 0.932  \\
    \textbf{\textit{EPRRNet(monocular)}} & \textbf{32.71} & \textbf{0.936}  \\
    \textit{PRRNet(stereo)} & 32.58 & 0.937  \\
    \textbf{\textit{EPRRNet(stereo)}} & \textbf{33.83} & \textbf{0.943}  \\
    \bottomrule
    \end{tabular}%
    \label{table_kitti2015}
\end{table}%

\begin{figure*}[tb]
  \centering
  \subfigure[Input]{
    \label{monocular:a}
    \includegraphics[width=0.18\linewidth ]{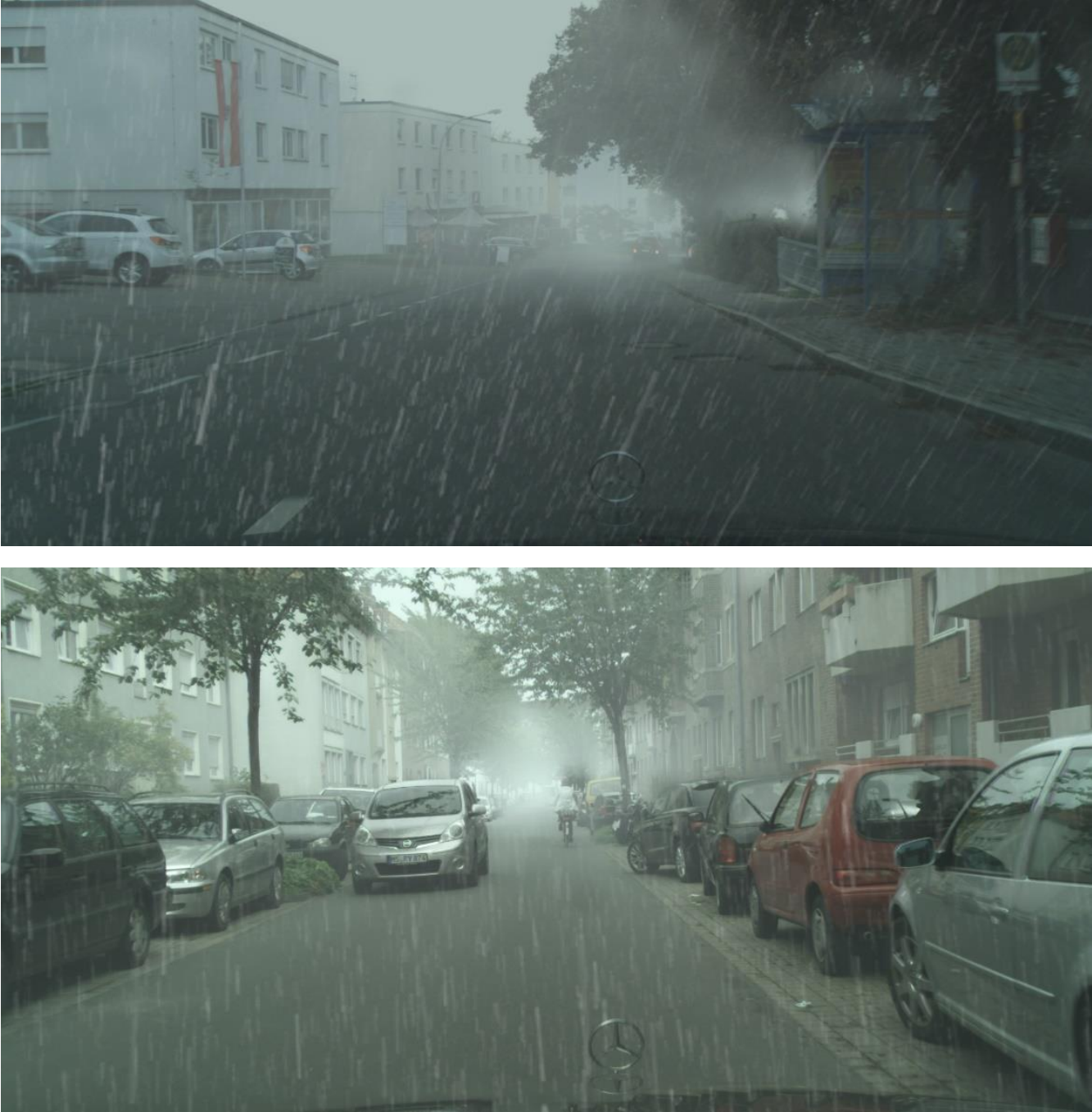}}
    \subfigure[DID-MDN]{
    \label{monocular:b}
    \includegraphics[width=0.18\linewidth ]{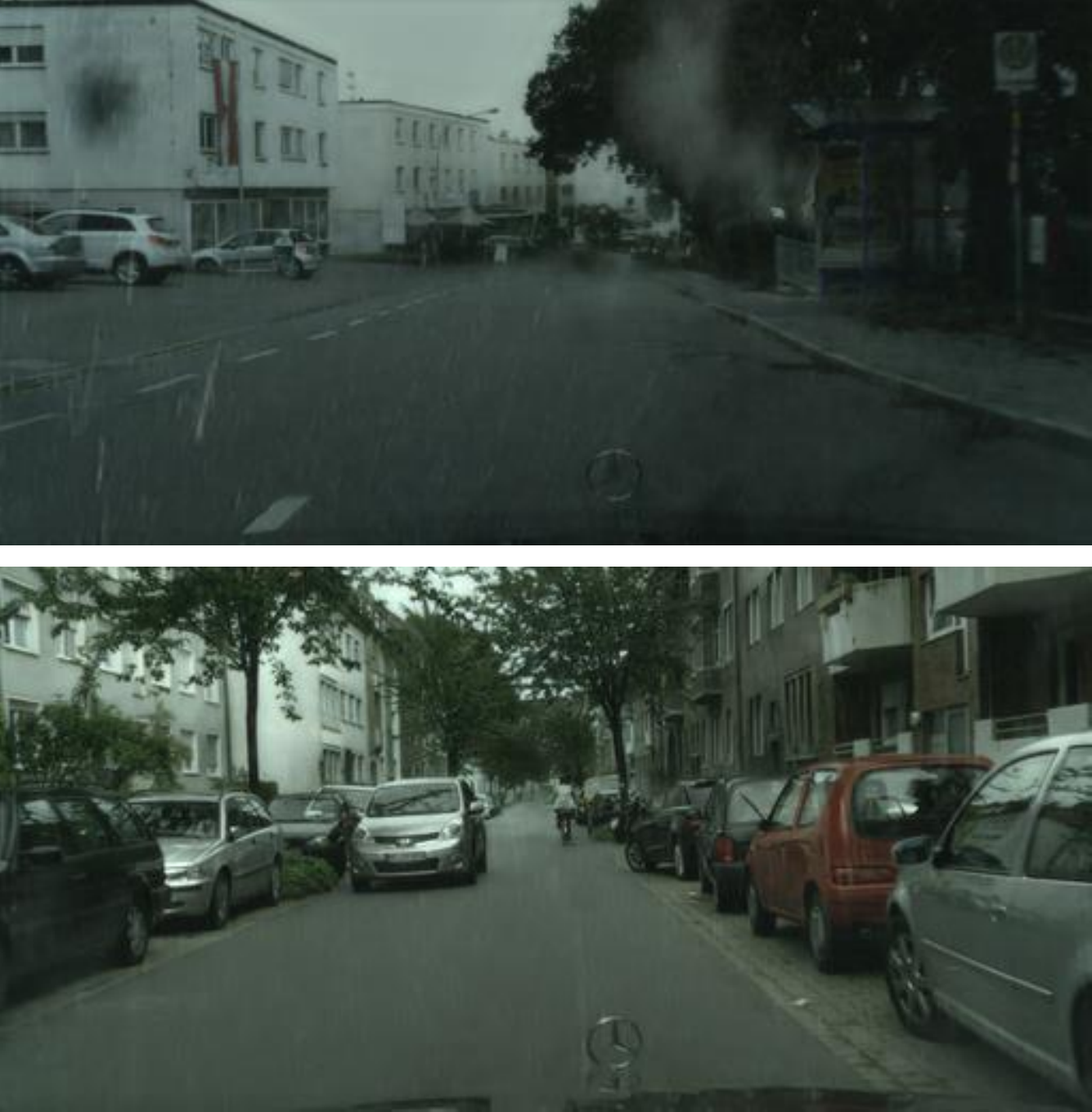}}
    \subfigure[DAF-Net]{
    \label{monocular:c}
    \includegraphics[width=0.18\linewidth ]{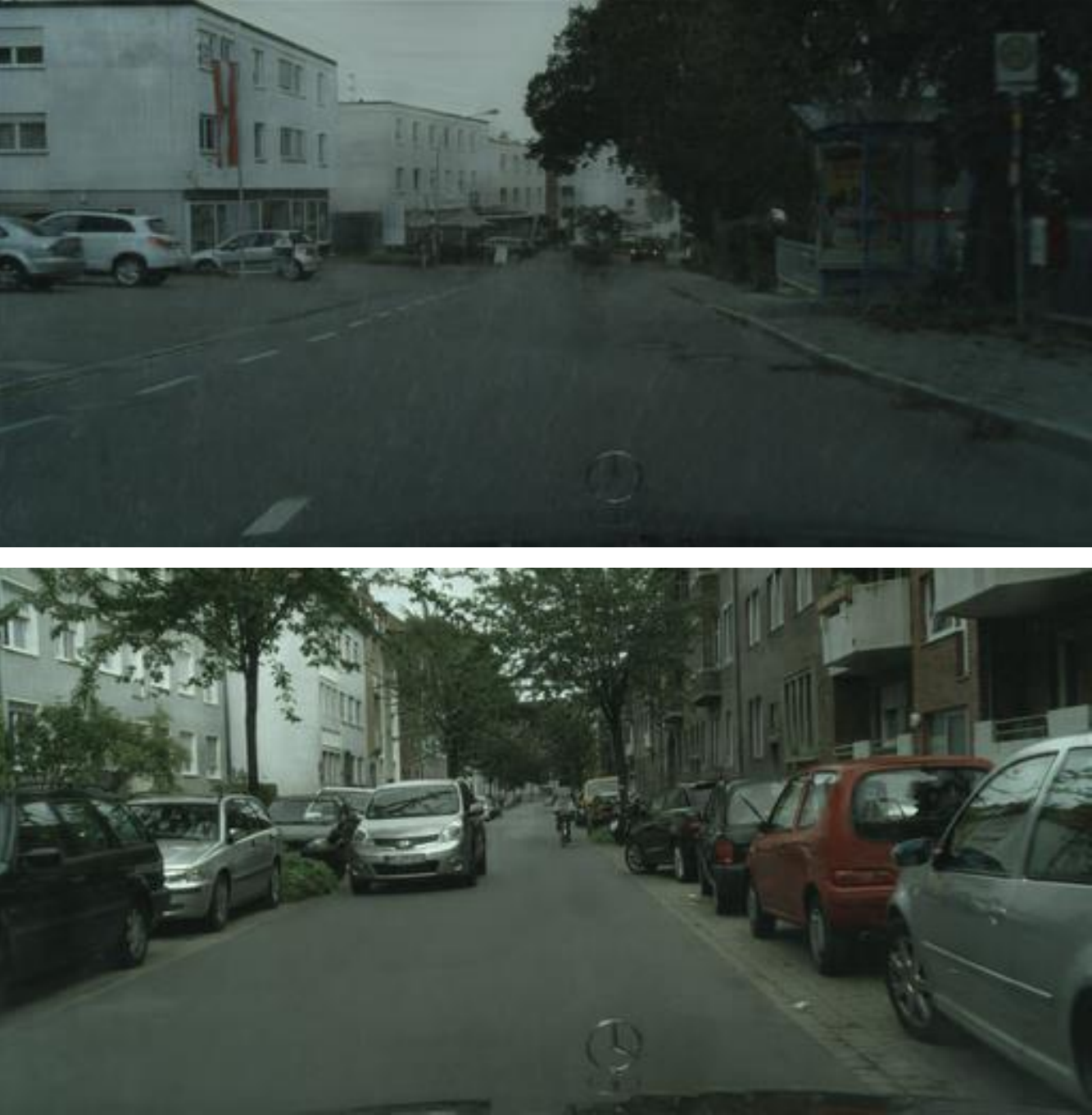}}
    \subfigure[Ours]{
    \label{monocular:d}
    \includegraphics[width=0.18\linewidth ]{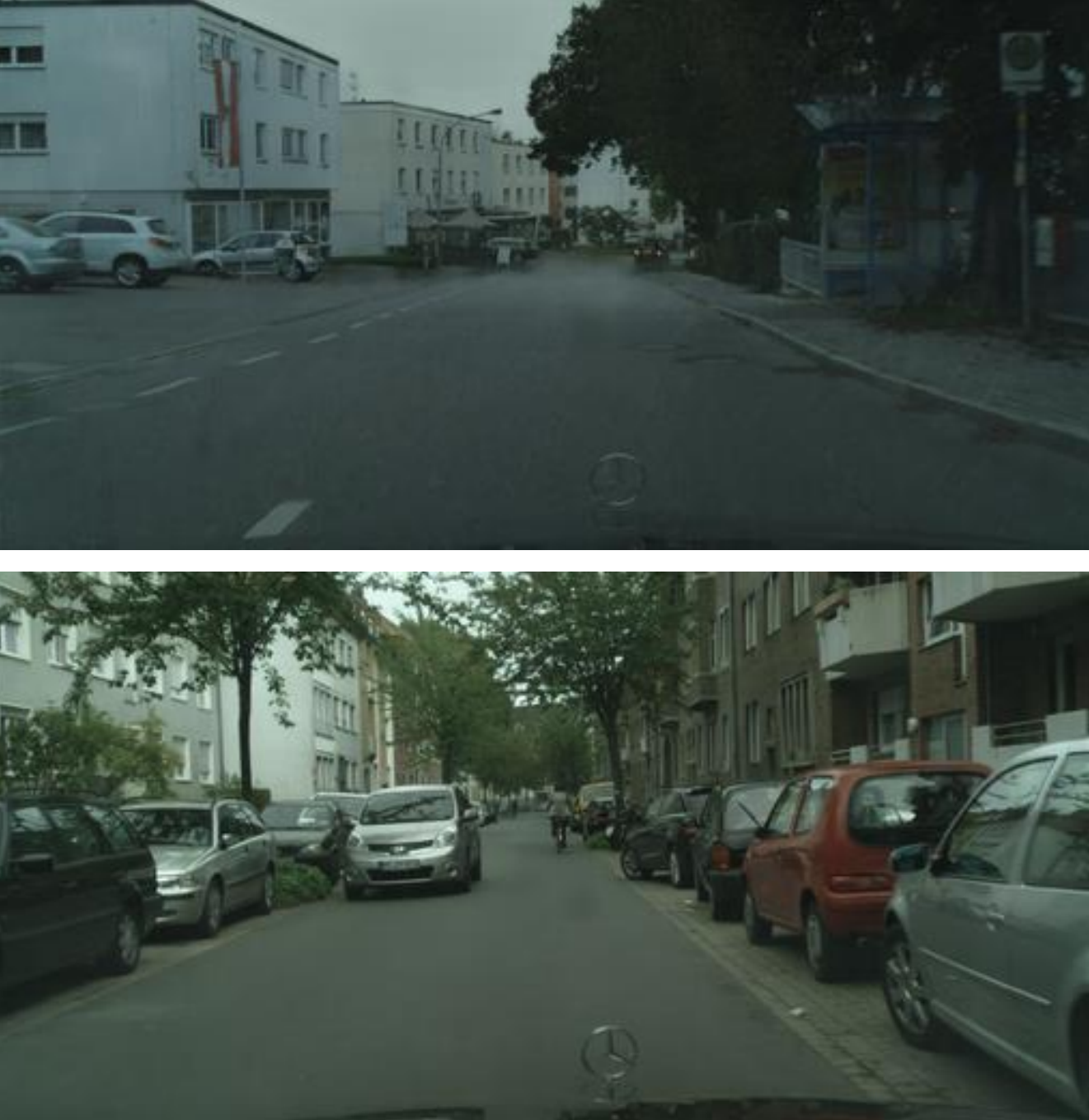}}
    \subfigure[Ground truth]{
    \label{monocular:d2}
    \includegraphics[width=0.18\linewidth ]{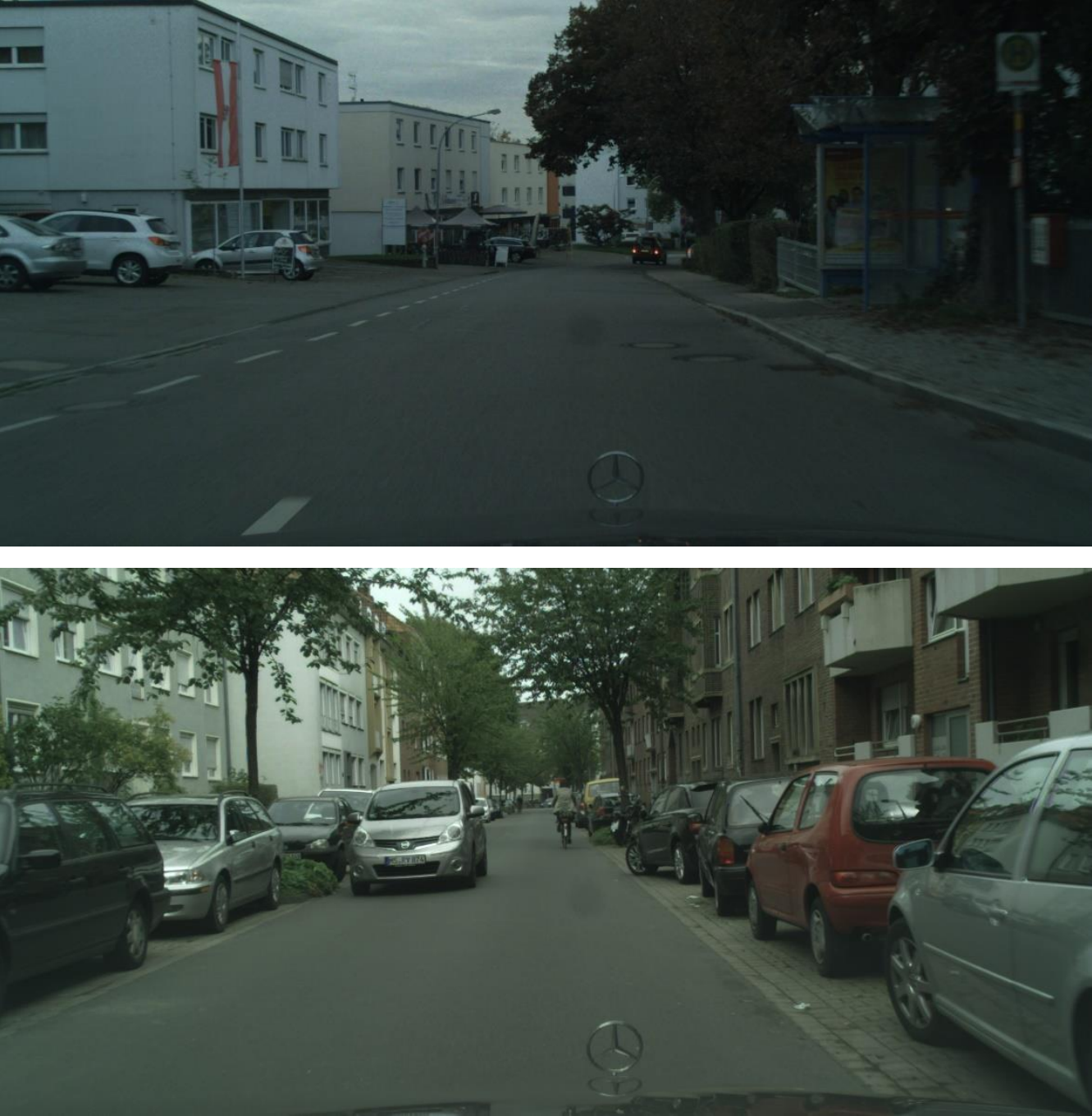}}
\caption{ \bf Exemplar results of qualitative evaluation of current state-of-the-art models on the RainCityscapes dataset.}
  \label{figure_city}
\end{figure*}

\begin{figure*}[tb]
  \centering
  \subfigure[{\scriptsize Input}]{
    \label{real:a}
    \includegraphics[width=0.185\linewidth ]{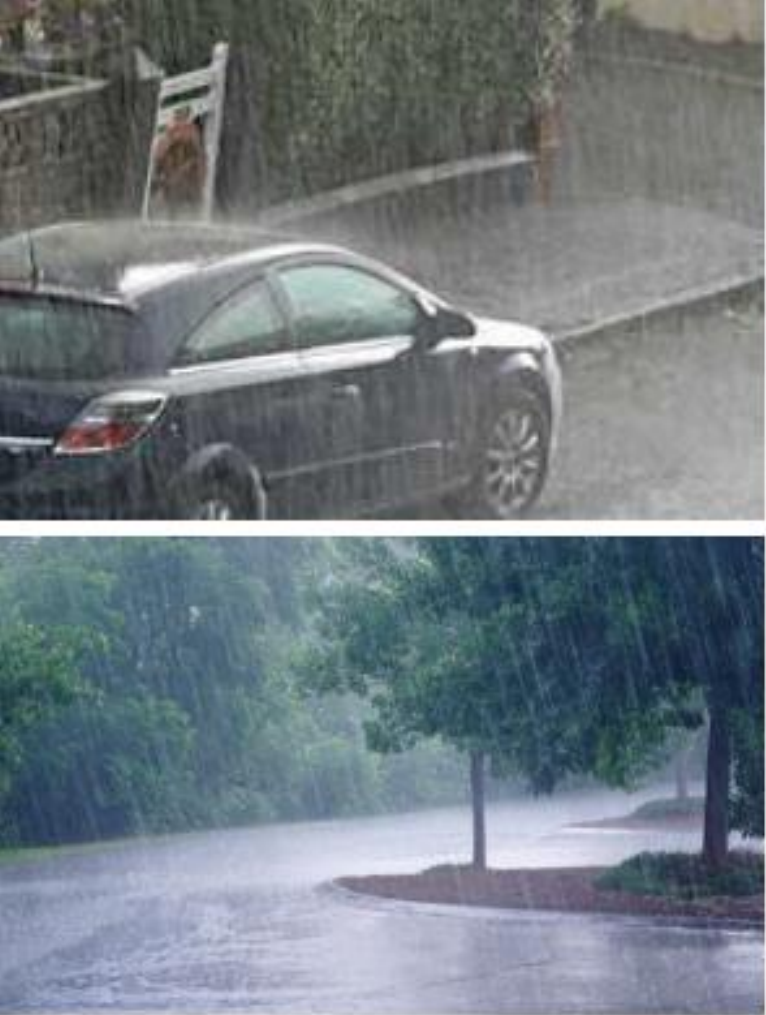}}
    \subfigure[{\scriptsize DAF-Net}]{
    \label{real:b}
    \includegraphics[width=0.185\linewidth ]{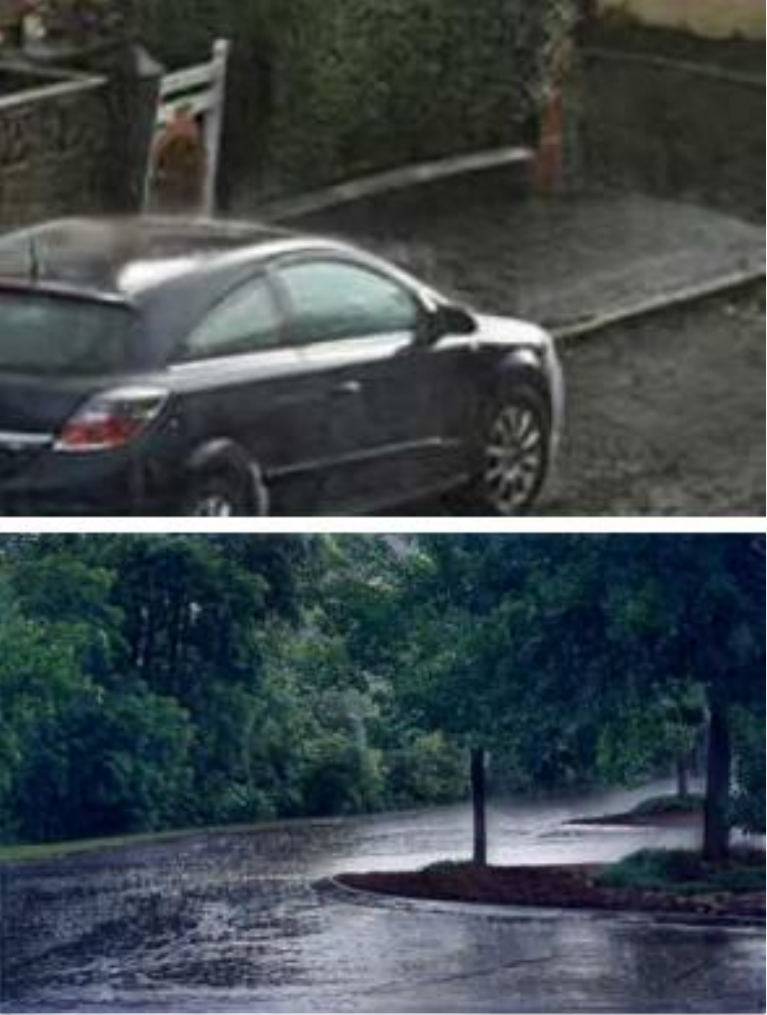}}
    \subfigure[{\scriptsize RESCAN}]{
    \label{real:c}
    \includegraphics[width=0.185\linewidth ]{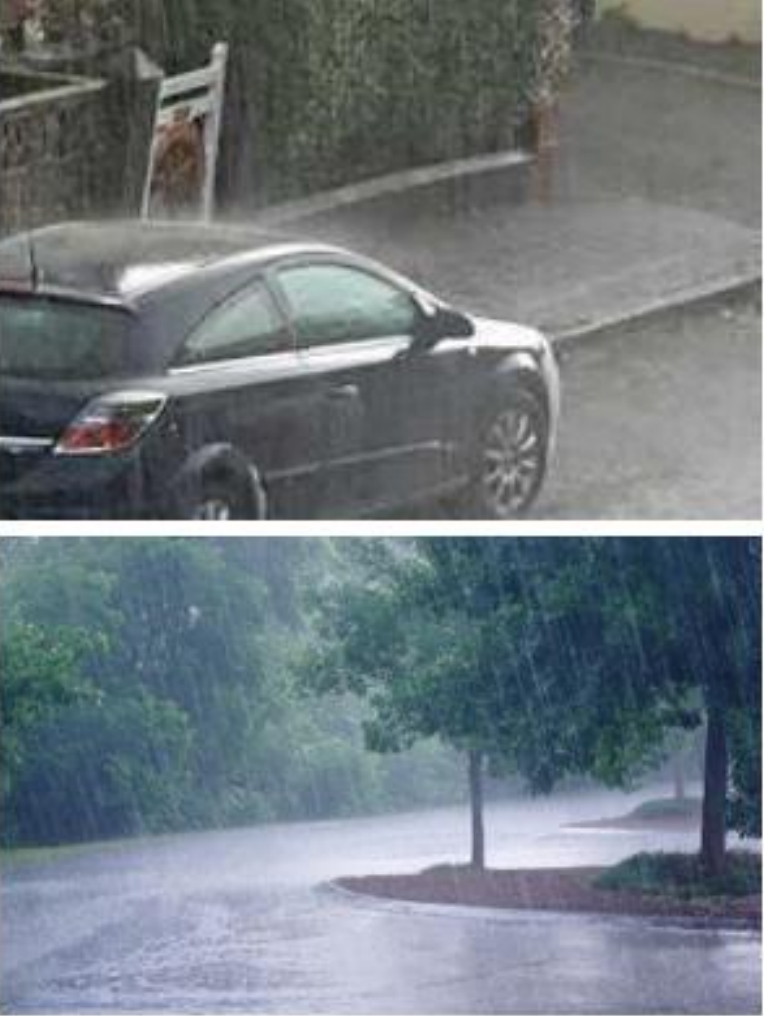}}
    \subfigure[{\scriptsize RESCAN + DCPDN}]{
    \label{real:d}
    \includegraphics[width=0.185\linewidth ]{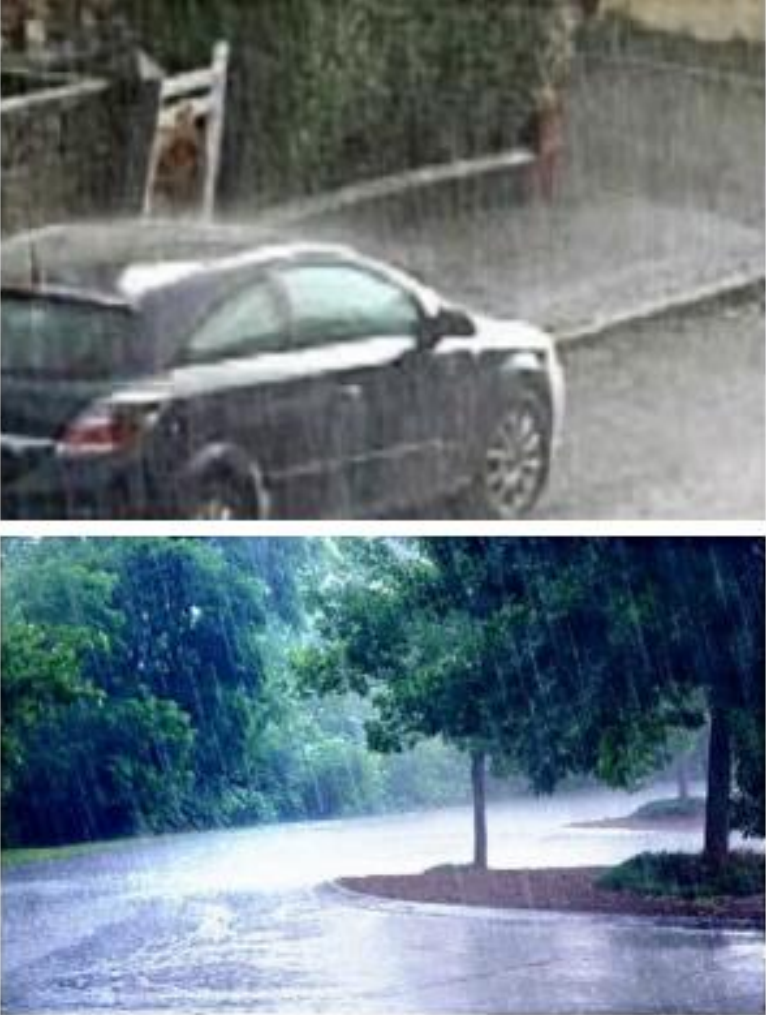}}
    \subfigure[{\scriptsize Ours}]{
    \label{real:d2}
    \includegraphics[width=0.185\linewidth ]{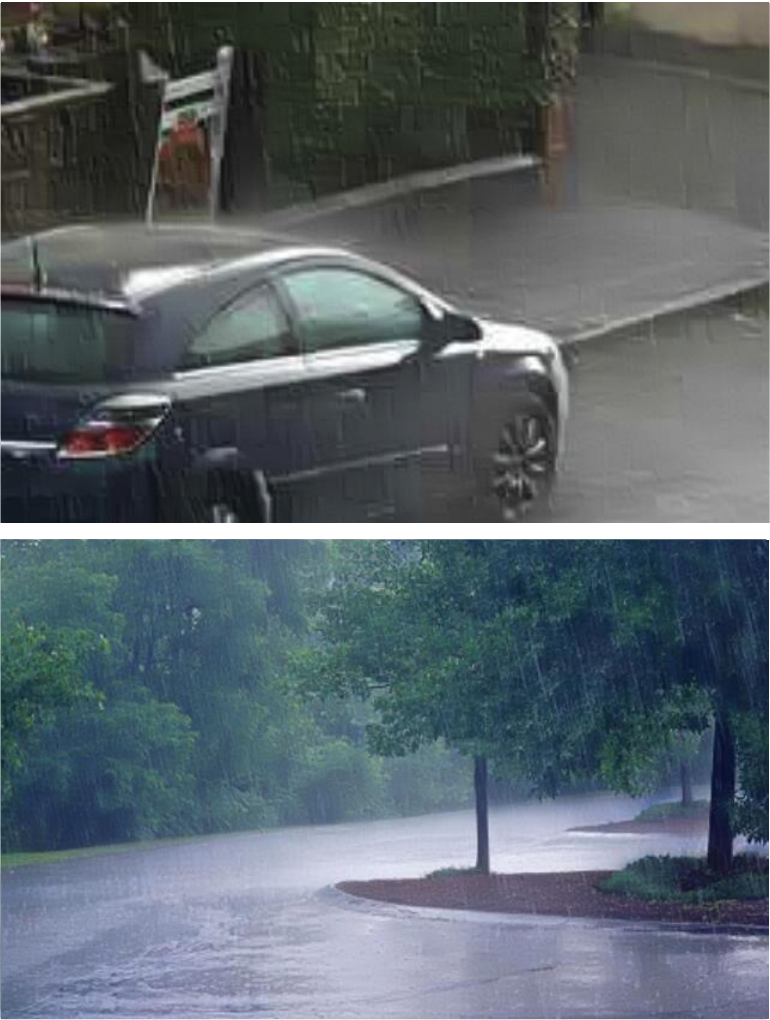}}
\caption{{\bf Qualitative evaluation on real-world rainy images.} From left to right are the input images, results of DAF-Net \cite{hu2019depth}, RESCAN \cite{li2018recurrent}, RESCAN + DCPDN \cite{zhang2018densely} and ours, respectively.}
  \label{figure_real}
\end{figure*}

\subsection{Monocular Deraining}
The proposed \textit{EPRRNet} is not only able to remove rain from stereo images, but also has the advantage of removing rain from a single image with its monocular version. In this section, we also evaluate it on the monocular dataset RainCityscapes.
We compare the monocular version of our models, \textit{PRRNet(monocular)} and \textit{EPRRNet(monocular)}, with the state-of-the-art methods, including DID-MDN \cite{zhang2018density}, RESCAN \cite{li2018recurrent}, JOB \cite{zhu2017joint}, GMMLP \cite{li2016rain}, DSC \cite{luo2015removing}, DCPDN \cite{zhang2018densely}, and DAF-Net \cite{hu2019depth}, from both quantitative and qualitative aspects.

The quantitative results on the RainCityscapes dataset are shown in Table \ref{table_city}. DID-MDN \cite{zhang2018density} and DCPDN \cite{zhang2018densely} perform well and DAF-Net \cite{hu2019depth} outperforms these two methods. Our monocular version \textit{EPRRNet(monocular)} achieves the best performance compared with all the compared methods on this task, revealing the effectiveness of taking semantic segmentation into consideration and the semantic-rethinking loop. 
Fig. \ref{figure_city} compares its qualitative performance with different methods. The results show that the monocular version of our \textit{EPRRNet} also achieves the best performance in terms of monocular image deraining.

\begin{table}[tb]
  \centering 
    \caption{\it Quantitative evaluation  of current state-of-the-art models on the RainCityscapes dataset.}
    \setlength\tabcolsep{5.0pt}
    \begin{tabular}{l |  c c }
    \toprule
    Methods &  PSNR & SSIM \\
    \hline
    DID-MDN \cite{zhang2018density} & 28.43 & 0.9349 \\
    RESCAN \cite{li2018recurrent} & 24.49 & 0.8852 \\
    JOB \cite{zhu2017joint} & 15.10 & 0.7592 \\
    GMMLP \cite{li2016rain} & 17.80 & 0.8169 \\
    DSC \cite{luo2015removing} & 16.25 & 0.7746 \\
    DCPDN \cite{zhang2018densely} & 28.52 & 0.9277 \\
    DAF-Net \cite{hu2019depth} & 30.06 & 0.9530 \\
    \hline
    \textbf{\textit{PRRNet}(monocular)} & \textbf{30.44} & \textbf{0.9688}  \\
    \textbf{\textit{EPRRNet}(monocular)} & \textbf{31.11} & \textbf{0.9741}  \\
    \bottomrule
    \end{tabular}%
    \label{table_city}
\end{table}%

\subsection{Evaluation on Real-world Images}

To further verify the effectiveness of our method, we show its performance of deraining on the real world rainy images. Fig. \ref{figure_real} shows the qualitative results on two exemplar images from the Internet. Compared to other competing methods, the proposed method \textit{EPRRNet(monocular)} achieves better performance via understanding the scene structure. For example, DAF-Net seems to generate well-derained images, but the produced derained images suffer from color distortion (\textit{e.g.}, the colors turn dark in the results). RESCAN and RESCAN+DCPDN perform worse than our method in removing rain.

\section{Conclusion}

In this paper, we present \textit{PRRNet}, the first stereo semantic-aware deraining network, for stereo image deraining. Different from previous methods which only learn from pixel-level loss functions or monocular information, the proposed model advances image deraining by leveraging semantic information extracted by a semantic-aware deraining model, as well as visual deviation between two views fused by two Fusion Nets, \textit{i.e.}, \textit{SFNet} and \textit{VFNet}.
We also synthesize two stereo deraining datasets to evaluate different deraining methods.
An enhanced version, \textit{i.e.}, \textit{EPRRNet} is developed with a parallel stereo fusion module.
The experimental results show that our proposed \textit{PRRNet} and \textit{EPRRNet} outperform the state-of-the-art methods on both monocular and stereo image deraining.

\section*{Acknowledgment}
This work is funded in part by the ARC Centre of Excellence for Robotics Vision (CE140100016), ARC-Discovery (DP 190102261) and ARC-LIEF (190100080) grants.  The authors gratefully acknowledge NVIDIA for GPU gift.


{\small
\bibliographystyle{spmpsci}
\bibliography{template}
}

\end{document}